\documentclass{article}

\usepackage[final,nonatbib]{neurips_2022}

\usepackage[utf8]{inputenc} 
\usepackage[T1]{fontenc}    
\usepackage{hyperref}       
\usepackage{url}            
\usepackage{booktabs}       
\usepackage{amsfonts}       
\usepackage{nicefrac}       
\usepackage{microtype}      
\usepackage{xcolor}         
\usepackage{mathtools}
\usepackage{multirow}
\usepackage{enumitem}[inline]
\usepackage{soul}
\usepackage{amssymb}
\usepackage{float}

\newtheorem{theorem}{Theorem}

\newtheorem{definition}{Definition}
\newtheorem{remark}{Remark}

\definecolor{ballblue}{rgb}{0.13, 0.67, 0.8}

\title{Learning Manifold Dimensions with Conditional Variational Autoencoders}

\author{%
  Yijia Zheng\textsuperscript{1}\thanks{Work completed during internship at the AWS Shanghai AI Labs. } \quad Tong He\textsuperscript{2} \quad Yixuan Qiu\textsuperscript{3} \quad David Wipf\textsuperscript{2}\\
  \textsuperscript{1} Department of Statistics, Purdue University \\
  \textsuperscript{2} Amazon Web Services \\
  \textsuperscript{3} School of Statistics and Management, Shanghai University of Finance and Economics\\
  \texttt{zheng709@purdue.edu, \{htong, daviwipf\}@amazon.com, qiuyixuan@sufe.edu.cn}
}

\begin{document}

\maketitle

\begin{abstract}
Although the variational autoencoder (VAE) and its conditional extension (CVAE) are capable of state-of-the-art results across multiple domains, their precise behavior is still not fully understood, particularly in the context of data (like images) that lie on or near a low-dimensional manifold. For example, while prior work has suggested that the globally optimal VAE solution can learn the correct manifold dimension, a necessary (but not sufficient) condition for producing samples from the true data distribution, this has never been rigorously proven.  Moreover, it remains unclear how such considerations would change when various types of conditioning variables are introduced, or when the data support is extended to a union of manifolds (e.g., as is likely the case for MNIST digits and related).  In this work, we address these points by first proving that VAE global minima are indeed capable of recovering the correct manifold dimension.  We then extend this result to more general CVAEs, demonstrating practical scenarios whereby the conditioning variables allow the model to adaptively learn manifolds of varying dimension across samples.  Our analyses, which have practical implications for various CVAE design choices, are also supported by numerical results on both synthetic and real-world datasets.  
\end{abstract}

\section{Introduction}\label{sec:intro}

Variational autoencoders (VAE)~\cite{Kingma2014, Rezende2014} and conditional variants (CVAE)~\cite{sohn2015learning} are powerful generative models that produce competitive results in various domains such as image synthesis~\cite{iclr/GulrajaniKATVVC17, razavi2019generating,van2017neural}, natural language processing~\cite{serban2017hierarchical}, time-series forecasting~\cite{pmlr-v177-lowe22a, tang2021probabilistic}, and trajectory prediction~\cite{li2021grin}. As a representative example, when equipped with an appropriate deep architecture, VAE models have recently achieved state-of-the-art performance generating large-scale images~\cite{peng2021generating}. And yet despite this success, there remain VAE/CVAE behaviors in certain regimes of interest where we lack a precise understanding or a supporting theoretical foundation.  

In particular, when the data lie on or near a low-dimensional manifold, as occurs with real-world images~\cite{pope2021intrinsic}, it is meaningful to have a model that learns the manifold dimension correctly.   The latter can provide insight into core properties of the data and  be viewed as a necessary, albeit not sufficient, condition for producing samples from the true distribution. Although it has been suggested in prior work~\cite{dai2021value,bin2019iclr} that a VAE model can learn the correct manifold dimension when globally optimized, this has only been formally established under the assumption that the decoder is linear or affine~\cite{dai2017hidden}.  And the potential ability to learn the correct manifold dimension becomes even more nuanced when conditioning variables are introduced. In this regard, a set of discrete conditions (e.g., MNIST image digit labels) may correspond with different ``slices'' through the data space, with each inducing a manifold with varying dimension (intuitively, the manifold dimension of images labelled ``1'' is likely smaller than those of ``5'').  Alternatively, it is possible to have data expand fully in the ambient space but lie on a low-dimensional manifold when continuous conditional variables are present. Such a situation can be trivially constructed by simply treating some data dimensions, or transformations thereof, as the conditioning variables.  In both scenarios, the role of CVAE models remains under-explored.

Moreover, unresolved CVAE properties in the face of low-dimensional data structure extend to practical design decisions as well.  For example, there has been ongoing investigation into the choice between a fixed VAE decoder variance and a learnable one~\cite{dai2021value,bin2019iclr,mattei2018leveraging,rezende2018taming,takahashi2018student}, an issue of heightened significance when conditioning variables are involved.  And there exists similar ambiguity regarding the commonly-adopted strategy of sharing weights between the prior and encoder/posterior in CVAEs~\cite{kipf2018neural, sohn2015learning}. Although perhaps not obvious at first glance, in both cases these considerations are inextricably linked to the capability of learning data manifold dimensions.


Against this backdrop our paper makes the following contributions:

\begin{enumerate}[label=(\roman*)]
    \item In Section \ref{sec:vae_manifold_dimensions} we provide the first demonstration of general conditions under which VAE global minimizers provably learn the correct data manifold dimension.  
    
    \item We then extend the above result in Section \ref{sec:cvae_manifold_dimensions} to address certain classes of CVAE models with either continuous or discrete conditioning variables, the latter being associated with data lying on a union of manifolds.

    
    \item Later, Section \ref{sec:model_design}  investigates common CVAE model designs and training practices, including the impact of strategies for handling the decoder variance as well as the impact of weight sharing between conditional prior and posterior networks.
    
    
    \item Section \ref{sec:experiments} supports our theoretical conclusions and analysis with numerical experiments on both synthetic and real-world datasets.
    
\end{enumerate}


\section{Learning the Dimension of Data Manifolds}\label{sec:manifold_learning}

In this section we begin with analysis that applies to regular VAE models with no conditioning.  We then later extend these results to more general CVAE scenarios.


\subsection{VAE Analysis} \label{sec:vae_manifold_dimensions}

We begin with observed variables $x \in \mathcal{X} \subseteq \mathbb{R}^d$, where $\mathcal{X}$ is the ambient data space equipped with some ground-truth probability measure $\omega_{gt}$.  Hence the probability mass of an infinitesimal $dx$ on $\mathcal{X}$ is $\omega_{gt}(dx)$ and $\int_{\mathcal{X}} \omega_{gt}(dx) = 1$.  VAE models attempt to approximate this measure with a parameterized distribution $p_\theta(x)$ instantiated as marginalization over latent variables $z \in \mathbb{R}^\kappa$ as in $p_\theta(x) = \int p_\theta(x|z) p(z)dz$.  Here $p(z) = N(z|0, I)$ is a standardized Gaussian prior and $p_\theta(x|z)$ represents a parameterized likelihood function that is typically referred to as the decoder.


To estimate decoder parameters $\theta$, the canonical VAE training loss is formed as a bound on the average negative log-likelihood given by
\begin{equation}\label{eq:elbo}
    \mathcal{L}(\theta, \phi) = \int_\mathcal{X} \{-\mathbb{E}_{q_\phi (z|x)}[\log p_\theta(x|z)] + \mathbb{KL}[q_\phi(z|x)||p(z)]\} \omega_{gt}(dx) \geq - \int_\mathcal{X} \log p_\theta(x) \omega_{gt}(dx),
\end{equation}
where the latent posterior distribution $q_\phi(z|x)$ (or the VAE encoder) controls the tightness of the bound via trainable parameters $\phi$.  Borrowing from~\cite{bin2019iclr}, we package widely-adopted VAE modeling assumptions into the following definition:


\begin{definition}[$\kappa$-simple VAE]
    A $\kappa$-simple VAE is a VAE model with dim[$z$] = $\kappa$ latent dimensions, the Gaussian encoder $q_\phi(z|x) = N(z| \mu_z(x; \phi), \mathrm{diag}\{\sigma_z^2(x;\phi)\})$, the Gaussian decoder $p_\theta(x|z) = N(x| \mu_x(z; \theta), \gamma I)$, and the prior $p(z) = N(z| 0, I)$. Here $\gamma > 0$ is a trainable scalar included within $\theta$, while the mean functions $\mu_z(x; \phi)$ and $\mu_x(z; \theta)$ are arbitrarily-complex $L$-Lipschitz continuous functions; the variance function $\sigma_z^2(x;\phi)$ can be arbitrarily complex with no further constraint.
\end{definition}

Our goal in this section will be to closely analyze the behavior of $\kappa$-simple VAE models when trained on data restricted to low-dimensional manifolds defined as follows:

\begin{definition}[Data lying on a manifold]
\label{data_def}
    Let $r$ and $d$ denote two positive integers with $r < d$.  Then $\mathcal{M}_r$ is a simple $r$-Riemannian manifold embedded in $\mathbb{R}^d$ when there exists a diffeomorphism $\varphi$ between $\mathcal{M}_r$ and $\mathbb{R}^r$. Specifically, for every $x \in \mathcal{M}_r$, there exists a $u = \varphi(x) \in \mathbb{R}^r$, where $\varphi$ is invertible and both $\varphi$ and $\varphi^{-1}$ are differentiable.
\end{definition}

As pointed out in \cite{bin2019iclr}, when training $\kappa$-simple VAEs on such manifold data, the optimal decoder variance will satisfy $\gamma \rightarrow 0$ (i.e, unbounded from below).  And as we will soon show, one effect of this phenomena can be to selectively push the encoder variances along certain dimensions of $z$ towards zero as well, ultimately allowing these dimensions to pass sufficient information about $x$ through the latent space such that the decoder can produce reconstructions with arbitrarily small error.  To formalize these claims, we require one additional definition:


\begin{definition}[Active VAE latent dimensions]\label{def:active_vae}
    Let $\{\theta^*_{\gamma}, \phi^*_{\gamma}\}$ denote globally-optimal parameters of a $\kappa$-simple VAE model applied to (\ref{eq:elbo}) as a function of an arbitrary fixed $\gamma$.  Then a dimension $j \in \{1,\ldots,\kappa \}$ of latent variable $z$ is defined as an \textit{active dimension} (associated with sample $x$) if the corresponding optimal encoder variance satisfies $\sigma_z(x; \phi^*_{\gamma})_j^2 \rightarrow 0$ as $\gamma \rightarrow 0$.
\end{definition}

We now arrive at the main result for this section:

\begin{theorem}[Learning the data manifold dimension using VAEs]\label{compression_vae}
     Suppose $\mathcal{X} = \mathcal{M}_r$ with $r < d$.  Then for all $\kappa \geq r$, any globally-optimal $\kappa$-simple VAE model applied to (\ref{eq:elbo}) satisfies the following:
     \begin{enumerate}
         \item[(i)] $\mathcal{L}(\theta^*_{\gamma}, \phi^*_{\gamma}) = (d - r) \log \gamma + O(1) $,
         \item[(ii)] The number of active latent dimensions almost surely equals $r$, and
         \item[(iii)] The reconstruction error almost surely satisfies $ \mathbb{E}_{q_{\phi_\gamma^\ast}(z|x)} \left[||x - \mu_x(z; \theta_\gamma^*)||^2\right] = O(\gamma)$.
         
     \end{enumerate}
\end{theorem}

While all proofs are deferred to the appendices, we provide a high-level sketch here. First, we prove by contradiction that there must exist at least $r$ active dimensions with corresponding encoder variances tending to zero at a rate of $O(\gamma)$.  If this is not the case, we show that the reconstruction term will grow at a rate of $O(\frac{1}{\gamma})$, leading to an overall loss that is unbounded from above. Next, we obtain upper bound and lower bounds on (\ref{eq:elbo}), both of which scale as $(d-r)\log\gamma + O(1)$ when the number of active dimensions is $r$. And lastly, we pin down the exact number of active dimensions by showing that the inclusion of any unnecessary active dimensions decreases the coefficient of the $\log\gamma$ scale factor, i.e., the factor $(d-r)$ uniquely achieves the minimal loss.

Overall, Theorem~\ref{compression_vae} provides a number of revealing insights into the VAE loss surface under the stated conditions.  First, we observe that although the loss can in principle be driven to minus infinity via sub-optimal solutions as $\gamma \rightarrow 0$, globally-optimal solutions nonetheless achieve an optimal rate (i.e., largest possible coefficient on the $\log \gamma$ factor) as well as the minimal number of active latent dimensions, which matches the ground-truth manifold dimension $r$.  Moreover, this all occurs while maintaining a reconstruction error that tends to zero as desired.  Additionally, while we have thus far treated $\gamma$ as a manually controlled parameter tending to zero for analysis purposes, when we transition to a trainable setting, similar intuitions are still applicable. More specifically, around global minimizers, the corresponding optimal $\gamma^*$ scales as $\frac{L^2}{d} || \sigma_z(x;\theta^*)_{1:r} ||^2$, where $\sigma_z(x;\theta^*)_{1:r}$ denotes the $r$ optimal latent posterior standard deviations associated with active dimensions.  Hence both the decoder variance $\gamma$ and the latent posterior variances of active dimensions converge to zero in tandem at the same rate; see the proof for more details.  In contrast, along the remaining inactive dimensions we show that $\lim_{\gamma \rightarrow 0} \sigma_z(x; \phi^*_{\gamma}) = \Omega(1)$, which optimzes the KL term without compromising the reconstruction accuracy.

In closing this section, we note that previous work \cite{bin2019iclr} has demonstrated that global minima of VAE models can achieve zero reconstruction error for all samples lying on a data manifold. But it was \textit{not} formally established in a general setting that this perfect reconstruction was possible using a \textit{minimal} number of active latent dimensions, and hence, it is conceivable for generated samples involving a larger number of active dimensions to stray from this manifold. In contrast, to achieve perfect reconstruction using the minimal number of active latent dimensions, as we have demonstrated here under the stated assumptions, implies that generated samples must also lie on the manifold. Critically, the noisy signals from inactive dimensions are blocked by the decoder and therefore cannot produce deviations from the manifold.

\subsection{Extension to Conditional VAEs} \label{sec:cvae_manifold_dimensions}




In this section, we extend our analysis to include conditional models, progressing from VAEs to CVAEs.  For this purpose, we introduce conditioning variables $c$ drawn from some set $\mathcal{C}$ with associated probability measure $\nu_{gt}$ such that $\int_\mathcal{C} \nu_{gt}(dc) = 1$.  Moreover, for any $c \in \mathcal{C}$ there exists a subset $\mathcal{X}_c \subseteq \mathcal{X}$ with probability measure $\omega_{gt}^c$ satisfying $\int_{\mathcal{X}_c} \omega^c_{gt}(dx) = 1$.  Collectively we also have $\int_{\mathcal{C}} \int_{\mathcal{X}_c} \omega^c_{gt}(dx)\nu_{gt}(dc) = 1$.  Given these definitions, the canonical CVAE loss is defined as


\begin{eqnarray}\label{eq:c_elbo}
     \mathcal{L}(\theta, \phi) & = & \int_\mathcal{C} \int_{\mathcal{X}_c} \left\{-\mathbb{E}_{q_\phi (z|x, c)}[\log p_\theta(x|z, c)] + \mathbb{KL}[q_\phi(z|x, c)||p_\theta(z| c)]\right\} \omega^c_{gt}(dx)\nu_{gt}(dc) \nonumber \\
     &\geq & \int_\mathcal{C} \int_{\mathcal{X}_c} -\log p_\theta(x|c) \omega^c_{gt}(dx)\nu_{gt}(dc),
\end{eqnarray}
which forms an upper bound on the conditional version of the expected negative log-likelihood.
We may then naturally extend the definition of the $\kappa$-simple VAE model to the conditional regime as follows:

\begin{definition}[$\kappa$-simple CVAE] \label{def:k_simple_cvae}
    A $\kappa$-simple CVAE is an extension of the $\kappa$-simple VAE with the revised conditional, parameterized prior $p_\theta(z| c) = N(z| \mu_z(c; \theta), \mathrm{diag}\{\sigma^2_z(c; \theta)\})$, the Gaussian encoder $q_\phi(z|x, c) = N(z| \mu_z(x, c; \phi), \mathrm{diag}\{\sigma^2_z(x, c; \phi)\})$, and the Gaussian decoder $p_\theta(x|z, c) = N(x| \mu_x(z, c; \theta), \gamma I)$. The encoder/decoder mean functions $\mu_z(x, c; \phi)$ and $\mu_x(z, c; \theta)$ are arbitrarily-complex $L$-Lipschitz continuous functions, while the prior mean $\mu_z(c; \theta)$ and variance $\sigma^2_z(c; \theta)$,\footnote{While the parameters of the prior mean and variance functions are labeled as $\theta$, this is merely to follow standard convention and group all parameters from the generative pipeline together under the same heading; it is not meant to imply that the decoder and prior actually share the same parameters.} and encoder variance $\sigma_z^2(x, c; \phi)$ can all be arbitrarily-complex functions with no further constraint.
\end{definition}


Likewise, we may also generalize the definition of active latent dimensions, where we must explicitly account for the modified conditional prior distribution which controls the relative scaling of the data.

\begin{definition}[Active CVAE latent dimensions]\label{def:active_cvae}
Let $\{\theta^*_{\gamma}, \phi^*_{\gamma}\}$ denote globally-optimal parameters of a $\kappa$-simple CVAE model applied to (\ref{eq:c_elbo}) as a function of an arbitrary fixed $\gamma$.  Then a dimension $j \in \{1,\ldots,\kappa \}$ of latent variable $z$ is defined as an \textit{active dimension} (associated with sample pair $\{x,c\}$) if the corresponding $j$-th optimal encoder/prior variance ratio satisfies $\sigma_{z}(x, c; \phi_\gamma^*)_j^2 / \sigma_{z}(c; \theta_{\gamma}^*)_j^2  \rightarrow 0$ as $\gamma \rightarrow 0$.
\end{definition}

Note that a CVAE (or VAE) with a standardized (parameter-free) Gaussian prior, the prior variance equals one.  Hence it follows that Definition~\ref{def:active_vae} is a special case of Definition~\ref{def:active_cvae} where $\sigma_z^2(c;\theta) = I$.

Before proceeding to our main result for this section, there is one additional nuance to our manifold assumptions underlying conditional data.  Specifically, if $c$ follows a continuous distribution, then it can reduce the number of active latent dimensions needed for obtaining perfect reconstructions.  To quantify this effect, let $\mathcal{M}_r^c \subseteq \mathcal{M}_r$ denote the subset of $\mathcal{M}_r$ associated with $\mathcal{X}_c$.  Intuitively then, depending on the information about $x$ contained in $c$, the number of active latent dimensions within $\mathcal{M}_r^c$ may be less than $r$.  We quantify this reduction via the following definition:

\begin{definition}[Effective dimension of a conditioning variable]
\label{def:effective_dim}
Given an integer $t \in \{0,\ldots,r \}$, let $\mathcal{C}_t$ denote a subset of $\mathcal{C}$ with the following properties: (i) There exists a function $g: \mathcal{C}_t \rightarrow \mathbb{R}^t$ as well as $t$ dimensions of $\varphi(x)$ denoted as $\varphi(x)_t$, such that $g(c) = \varphi(x)_t$ for all pairs $\{ (c, x) : c \in \mathcal{C}_t,~~x \in \mathcal{M}_r^c \}$, where $\varphi$ is a diffeomorphism per Definition~\ref{data_def}; and (ii)  there does not exist such a function $g$ for $t+1$.  We refer to $t$ as the effective dimension of any conditioning variable $c \in \mathcal{C}_t$.
\end{definition}

Loosely speaking, Definition~\ref{def:effective_dim} indicates that any $c \in \mathcal{C}_t$ can effectively be used to reconstruct $t \leq r$ dimensions of $x$ within $\mathcal{M}_r^c$.  Incidentally, if $t = r$, then this definition implies that $x$ degenerates to a deterministic function of $c$.  Given these considerations, Theorem~\ref{compression_vae} can be extended to CVAE models conditioned on continuous variables as follows:

\begin{theorem}[Learning the data manifold dimension using CVAEs]\label{compression_cvae}
      Suppose $\mathcal{C} = \mathcal{C}_t$ and $\mathcal{X}_c = \mathcal{M}_r^c$ with $r \geq 1$, $t \geq 0$, and $r \geq t$.
       Then any globally-optimal $\kappa$-simple CVAE model applied to the loss (\ref{eq:c_elbo}), with $\kappa \geq r$, satisfies the following:
\begin{enumerate}
     \item[(i)] $\mathcal{L}(\theta^*_{\gamma}, \phi^*_{\gamma}) = (d - r + t) \log \gamma + O(1)$, and
     \item[(ii)] The number of active latent dimensions almost surely equals $r-t$, and
     \item[(iii)] The reconstruction error almost surely satisfies $\mathbb{E}_{q_{\phi_\gamma^\ast}(z|x, c)} \left[||x - \mu_x(z, c; \theta_\gamma^*)||^2\right]  = O(\gamma)$.
 \end{enumerate}
\end{theorem}

This result indicates that conditioning variables can further reduce the CVAE loss (by increasing the coefficient on the $\log \gamma$ term as $\gamma \rightarrow 0$ around optimal solutions.  Moreover, conditioning can replace active latent dimensions; intuitively this occurs because using $c$ to reconstruct dimensions of $x$, unlike dimensions of $z$, incurs no cost via the KL penalty term.  Additionally, it is worth mentioning that even if the observed data itself is not strictly on a manifold (meaning $r = d$), once the conditioning variables $c$ are introduced, manifold structure can be induced on $\mathcal{X}_c$, i.e., with $t>0$ it follows that $d-r+t > 0$ and the number of active latent dimensions satisfies $r-t < d$.



\subsection{Adaptive Active Latent Dimensions}
\label{sec:adaptive_dim}

Thus far our analysis has been predicated on the existence of a single $r$-dimensional manifold underlying the data $x$, along with a conditioning variable $c$ that captures $t$ degrees-of-freedom within this manifold.  More broadly though, it is reasonable to envision scenarios whereby the data instead lie on a union of manifolds, each with a locally-defined value of $r$ and possibly $t$ for continuous conditioning variables.  In such instances, both Theorems \ref{compression_vae} and \ref{compression_cvae} can be naturally refined to reflect this additional flexibility.  

While we defer a formal treatment to future work, it is nonetheless worthwhile to consider CVAE behavior within two representative scenarios. First, consider the case where $c$ is now a \textit{discrete} random variable taking a value in some set/alphabet $\{ \alpha_k \}_{k=1}^m$, such as the label of an MNIST digit \cite{mnist} (whereby $m = 10$).  It then becomes plausible to assume that $r = f(\alpha_k)$ for some function $f$, meaning that the manifold dimension itself may vary conditioned on the value of $c$ (e.g., the space of digits ``1" is arguably simpler than the space of digits ``8").  In principle then, a suitably-designed CVAE model trained on such data should be able to adaptively learn the dimensions of these regional manifolds.  Later in Section \ref{sec:adaptive_active_dims} we empirically demonstrate that when we include a specialized attention layer within the CVAE decoder, which allows it to selectively shut on and off different latent dimensions, the resulting model can indeed learn the underlying union of low-dimensional manifolds under certain circumstances.  From a conceptual standpoint, the outcome is loosely analogous to a separate, class-conditional VAE being trained on data associated with each $\alpha_k$.

As a second scenario, we may also consider the case where $t$ varies for different values of a \textit{continuous} conditioning variable $c$. Extrapolating from Theorem \ref{compression_cvae}, we would expect that the number of active latent dimensions $r-t$ will now vary across regions of the data space.  Hence an appropraite CVAE architecture should be able to adaptively compress the active latent dimensions so as to align with the varying information contained in $c$.  Again, Section \ref{sec:adaptive_active_dims} demonstrates that this is indeed possible.

\section{On Common CVAE Model Design Choices}
\label{sec:model_design}

In this section, we review CVAE model designs and training practices that, while adopted in various prior CVAE use cases, nonetheless may have underappreciated consequences, especially within the present context of learning the underlying data manifold dimensionality. 


\subsection{On the Equivalence of Conditional and Unconditional Priors} \label{sec:equivalency_of_priors}

Per the canonical CVAE design, it is common to include a parameterized, trainable prior $p_\theta(z|c)$
within CVAE architectures~\cite{bauer2019resampled,kipf2018neural, li2021grin, tomczak2018vae, zhu2017toward}. However, the strict necessity of doing so is at least partially compromised by the following remark:



\begin{remark}[Converting conditional to unconditional priors]
\label{rmk:equal_prior}
Consider a $\kappa$-simple CVAE model with prior $p_\theta(z|c)$, encoder $q_\phi(z|x, c)$ and decoder $p_\theta(x|z, c)$. We can always find another $\kappa$-simple CVAE model with prior $p(z)\sim\mathcal{N}(0, I)$, encoder $q_{\phi^\prime}(z|x, c)$ and decoder $p_{\theta^\prime}(x|z, c)$, such that $\mathcal{L}(\theta, \phi)$ = $\mathcal{L}({\theta^\prime}, {\phi^\prime})$ and $p_\theta(x|c) = p_{\theta'}(x|c)$.

\end{remark}

Remark~\ref{rmk:equal_prior} indicates that, at least in principle, a parameterized, conditional prior is not unequivocally needed. Specifically, as detailed in the appendix, we can always explicitly convert an existing $\kappa$-simple CVAE model with conditional prior $p_\theta(z|c)$ into another $\kappa$-simple CVAE model with fixed prior $p(z) = \mathcal{N}(z|0, I)$ without sacrificing any model capacity in the resulting generative process $p_{\theta'}(x|c) = \int p_{\theta'}(x|z,c) p(z) dz$; essentially the additional expressivity of the conditional prior is merely absorbed into a revised decoder.  Even so, there may nonetheless remain differences in the optimization trajectories followed during training such that achievable local minima may at times lack this equivalence.



\subsection{The Impact of $\gamma$ Initialization on Model Convergence}\label{subsec:gamma}

As emphasized previously, VAE/CVAE models with sufficient capacity applied to manifold data achieve the minimizing loss when $\gamma \rightarrow 0$.  However, directly fixing $\gamma$ near zero can be problematic for reasons discussed in \cite{dai2021value}, and more broadly, performance can actually be compromised when $\gamma$ is set to any fixed positive constant as noted in~\cite{bin2019iclr, mattei2018leveraging, rezende2018taming, takahashi2018student}.

Even so, it remains less well-understood how the initializaton of a learnable $\gamma$ may impact the optimization trajectory during training.  After all, the analysis we have provided is conditioned on finding global solutions, and yet it is conceivable that different $\gamma$ initializations could influence a model's ability to steer around bad local optima.  Note that the value of $\gamma$ at the beginning of training arbitrates the initial trade-off between the reconstruction and KL terms, as well as the smoothness of the initial loss landscape, both of which are factors capable of influencing model convergence.  We empirically study these factors in Section~\ref{subsec:init_gamma_exp}.

\subsection{A Problematic Aspect of Encoder/Prior Model Weight Sharing}








Presumably to stabilize training and/or avoid overfitting, one widely-adopted practice is to share CVAE weights between the prior and posterior/encoder modules \cite{li2021grin, sohn2015learning, tang2021probabilistic}. For generic, fixed-sized input data this may take the form of simply constraining the encoder as $q_\phi(z|x,c) = p_\theta(z|c)$  \cite{sohn2015learning}.  More commonly though, for sequential data both prior and encoder are instantiated as some form of recurrent network with shared parameters, where the only difference is the length of the input sequence \cite{li2021grin, tang2021probabilistic}, i.e., the full sequence for the encoder or the partial conditioning sequence for the prior.

More concretely with respect to the latter, assume sequential data $x = \{x_l\}_{l=1}^n$, where $l$ is a time index (for simplicity here we will assume a fixed length $n$ across all samples, although this can be easily generalized).  Then associated with each time point $x_l$ within a sample $x$, we have a prior conditioning sequence $c_l = x_{<l} \triangleq \{x_j\}_{j=1}^{l-1}$.  The resulting encoder and prior both define distributions over a corresponding latent $z_l$ via $q_\phi (z_l|x_{\leq l})$ and $p_\theta(z_l| x_{< l})$ respectively.  Along with the decoder model $p_\theta(x_l|z_l, x_{< l})$, the revised, sequential CVAE training loss becomes
\begin{equation} \label{eq:seq_cvae_loss}
        \mathcal{L}(\theta, \phi) = \int_\mathcal{X} \sum_{l=1}^n \{-\mathbb{E}_{q_\phi (z_l|x_{\leq l})}\left[\log p_\theta(x_l|z_l, x_{< l}\right] + \mathbb{KL}\left[q_\phi(z_l|x_{\leq l})||p_\theta(z| x_{< l})\right]\} \omega_{gt}(d x),
\end{equation}
where we observe that there is now an additional summation over the temporal index $l$. This mirrors the fact that to reconstruct (or at test time generate) a sample $x$, the CVAE will sequentially produce each time point $x_l$ conditioned on previously reconstructed (or generated) values. We now analyze one of the underappreciated consequences of encoder/prior weight-sharing within the aforementioned sequential CVAE context.

\begin{theorem}[Weight sharing can compromise the performance of sequential CVAEs]\label{thm:share_weight}
    Assume sequential data with ground-truth measure $\omega_{gt}$ defined such that the probability mass of $x_l$ conditioned on $x_{<l}$ lies on a manifold with minimum dimension $r>1$ (this excludes strictly deterministic sequences). Moreover, given a $\kappa$-simple sequential CVAE model,\footnote{This is defined analogously to the $\kappa$-simple CVAE from Definition \ref{def:k_simple_cvae}.} assume that the prior is constrained to share weights with the encoder such that $p_\theta(z_l|x_{<l}) = q_\phi(z_l|x_{< l})$.  Then the corresponding loss from (\ref{eq:seq_cvae_loss}) satisfies $\mathcal{L}(\theta, \phi) = \Omega(1)$ for any $\theta$ and $\phi$.
    
    
\end{theorem}
Given the rigidity of this lower bound, Theorem \ref{thm:share_weight} indicates that the sequential CVAE model is unable to drive the the loss towards minus infinity as required to produce active latent dimensions (per Definition \ref{def:active_cvae}) and correctly learn low-dimensional manifolds when present.  Intuitively, this relates to subtleties of the differing roles the encoder and prior play in sequence models.  

In particular, it follows that the encoder distribution $q_\phi(z_l|x_{\leq 1})$ at time $l$ is in every way \textit{identical} to the prior at time $l+1$ given that $p_\theta(z_{l+1}| x_{<l+1}) = q_\phi(z_{l+1}| x_{<l+1}) = q_\phi(z_{l+1}| x_{\leq l})$ per the adopted weight sharing assumption.  But these distributions are meant to serve two very different purposes: (i) The \textit{encoder} is meant to push the variances of active dimensions to zero so as to learn the underlying manifold, while the remaining/superfluous dimensions merely output useless noise that is filtered by the decoder. (ii) In stark contrast, the role of the \textit{prior} is not to instantiate reconstructions, but rather to inject the calibrated randomness needed to match the conditional uncertainty in the ground-truth distribution.  Therefore, unless the observed data sequences are deterministic, in which case $x_l$ can be predicted exactly from $x_{<l}$, it is \textit{impossible} to achieve both (i) and (ii) simultaneously.  And indeed, we verify these insights via numerical experiments conducted in  Section~\ref{subsec:share_weight_exp}.

\vspace*{-0.1cm}
\section{Experiments}
\label{sec:experiments}
\vspace*{-0.1cm}
In this section we first corroborate our previous analysis in a controllable, synthetic environment; later we extend to real-world datasets to further support our conclusions.\footnote{Code is available at \href{https://github.com/zhengyjzoe/manifold-dimensions-cvae}{https://github.com/zhengyjzoe/manifold-dimensions-cvae}} We also include experiments comparing the impact of full versus diagonal encoder covariance models to validate with respect to complementary analysis of VAE \textit{local} minima from \cite{wipf2023}. Note that the VAE/CVAE  model architectures used for all of our experiments are described in Section~A of the appendix.

\vspace*{-0.1cm}
\subsection{Datasets and Metrics}
\vspace*{-0.1cm}

\paragraph{Synthetic Data}

We begin by generating low-dimensional data $u \in \mathbb{R}^r$ from a Gaussian mixture model with 5 equiprobability components, each with mean $\mu_u$ and transformed variance $\log \sigma_u$ drawn from a standardized Gaussian distribution.  Next, to produce the high-dimensional data manifold $\mathcal{X} = \mathcal{M}_r$ in $\mathbb{R}^d$, we compute $x = \text{sigmoid}(G_x u)$, where $G_x \in \mathbb{R}^{d \times r}$ is a randomly initialized projection matrix.  For conditional models, we also compute  $c = G_c u_{1:t}$, where $G_c \in \mathbb{R}^{t \times t}$ is another random projection.  Given these assumptions, we can then construct training datasets with varying values of $r$, $d$, and $t$.
For all experiments with synthetic data, the training set size is 100,000. We also perform tests using synthetic sequential data described in Section \ref{subsec:share_weight_exp}.


\paragraph{Real-world Data}

We also investigate model behaviors via MNIST \cite{mnist} and Fashion MNIST~\cite{xiao2017fashion}.\footnote{MNIST is under the CC BY-SA 3.0 license, and Fashion MNIST is MIT-Licensed.} These two datasets involve image samples with clear-cut low-dimensional manifold structure, while at the same time, they are not so complex that more intricate architectures and training designs are required that might otherwise obfuscate our intended message (e.g., more complex models may fail at times to learn correct low-dimensional structure simply because of convergence issues).


\paragraph{Metrics}

By convention, we report the loss from (\ref{eq:elbo}) for VAEs and (\ref{eq:c_elbo}) for CVAEs; these losses are often referred to as the negative evidence lower bound (-ELBO) (note though that these values are not always directly comparable across different testing scenarios). 
We also include auxiliary metrics to diagnose model behavior, including  the number of active dimensions (AD), the reconstruction error (Recon), the KL-divergence ($\mathbb{KL}$), and the learned decoder variance $\gamma$.

\subsection{Learning Manifold Dimensions with VAEs}

We begin by exploring the VAE's ability to learn the correct manifold dimension in support of Theorem~\ref{compression_vae}.  Specifically, using the synthetic dataset, we vary the ambient dimension $d \in \{10,20,30\}$ and the ground-truth manifold dimension $r  \in \{2,4,6,8,10 \}$ and compare with the estimated number of active dimensions with $\kappa =\in \{5,20\}$.  Please see Section B of the appendix for details regarding how active dimensions are computed in practice (while in numerical experiments encoder variances will not converge to exactly zero, we can nonetheless observe these variances closely clustering around either 0 or 1 as expected).

Results are shown in Table~\ref{tab:vae_latent}, where we observe that when $\kappa \geq r$, the VAE can generally align the active dimensions with the value of $r$ as expected once $\gamma$ has converged to a small value.  In contrast, when $\kappa$ is too small (i.e., $\kappa = 5$), it is no longer possible to learn the manifold dimension (not surprisingly, the -ELBO and reconstruction error are also much larger as well).


\begin{table}[]
\centering
\caption{Aligning learned VAE active dimensions (AD) with the ground-truth manifold dimension $r$. When $\kappa \geq r$ (a surplus of latent dimensions) the VAE largely succeeds as $\gamma$, the reconstruction error (Recon), and -ELBO converge to relatively small values; however, when $\kappa < r$ this is not possible.}
    \label{tab:vae_latent}
\begin{tabular}{cccccccc}
\toprule
$\kappa$            & $d$                 & $r$ & AD & Recon & $\mathbb{KL}$ & $\gamma$            & -ELBO     \\ \midrule
\multirow{16}{*}{20} & \multirow{5}{*}{10} & 2   & 2                & 3$\times 10^{-4}$              & 18.31         & 1.625$\times 10^{-5}$ & -58.26  \\
                    &                     & 4   & 4                & 2.6$\times 10^{-3}$            & 24.22         & 5.654$\times 10^{-5}$ & -29.83  \\
                    &                     & 6   & 6                & 9.2$\times 10^{-3}$            & 24.14         & 3$\times 10^{-4}$     & -17.39  \\ 
                    &                     & 8   & 7                & 1.27$\times 10^{-2}$           & 27.91         & 1.4$\times 10^{-3}$   & -10.38  \\
                    &                     & 10  & 8                & 5.99$\times 10^{-2}$           & 16.39         & 2.5$\times 10^{-3}$   & -6.40   \\ \cmidrule(r){2-8}
 & \multirow{5}{*}{20} & 2   & 2                & 1.6$\times 10^{-3}$            & 17.98         & 5.052$\times 10^{-5}$ & -114.52 \\
                   &                     & 4   & 4                & 1.75$\times 10^{-2}$           & 23.11         & 2$\times 10^{-4}$     & -60.90  \\
                    &                     & 6   & 6                & 3.09$\times 10^{-2}$           & 28.96         & 6$\times 10^{-4}$     & -43.75  \\
                    &                     & 8   & 8                & 3.42$\times 10^{-2}$           & 33.83         & 1.2$\times 10^{-3}$   & -36.82  \\
                    &                     & 10  & 10               & 4.74$\times 10^{-2}$           & 35.81         & 1.1$\times 10^{-3}$   & -28.34  \\ \cmidrule(r){2-8}
 & \multirow{5}{*}{30} & 2   & 2                & 2.6$\times 10^{-3}$            & 18.42         & 7.221$\times 10^{-5}$ & -176.74 \\
                    &                     & 4   & 4                & 2.73$\times 10^{-2}$           & 24.60         & 2$\times 10^{-4}$     & -100.28 \\
                    &                     & 6   & 6                & 4.74$\times 10^{-2}$           & 31.89         & 9$\times 10^{-4}$     & -76.46  \\
                    &                     & 8   & 8                & 5.68$\times 10^{-2}$           & 37.28         & 1.6$\times 10^{-3}$   & -65.66  \\
                    &                     & 10  & 10                & 1.13$\times 10^{-1}$           & 35.13         & 2.5$\times 10^{-3}$   & -47.00  \\ \cmidrule(r){1-1} \cmidrule(r){2-8}
\multirow{3}{*}{5}  & \multirow{3}{*}{20} & 6   & 5                & 1.299$\times 10^{-1}$          & 22.53         & 2.1$\times 10^{-3}$   & -36.97  \\
                    &                     & 8   & 5                & 3.719$\times 10^{-1}$          & 16.618        & 8.8$\times 10^{-3}$   & -22.60  \\
                    &                     & 10  & 5                & 3.564$\times 10^{-1}$          & 15.966        & 1.113$\times 10^{-2}$ & -16.96  \\ \bottomrule
\end{tabular}

\end{table}


Complementary results on MNIST and Fashion MNIST are reported in Table~\ref{tab:vae_latent_mnist}. While we no longer have access to the ground-truth value of $r$, we can still observe that as $\kappa$ increases, the number of active dimensions saturates as expected. 



\begin{table}[]
    \centering
\caption{VAE latent compression on real datasets in further support of Theorem~\ref{compression_vae}.}\label{tab:vae_latent_mnist}
\begin{tabular}{ccccc}
    \toprule
                    Dataset           & $\kappa$ & AD & Recon & -ELBO     \\ \cmidrule(r){1-1} \cmidrule(r){2-5}
\multirow{3}{*}{MNIST}         & 5        & 5                & 14.899                         & -842.286  \\
                               & 16       & 12               & 9.749                          & -1065.83 \\
                               & 32       & 13               & 7.469                          & -1224.37 \\
                               \cmidrule(r){1-1} \cmidrule(r){2-5}
\multirow{3}{*}{Fashion MNIST} & 5        & 5                & 13.163                         & -935.127 \\
                               & 16       & 9               & 9.026                          & -1216.68 \\
                               & 32       & 9               & 7.820                          & -1327.26 \\
    \bottomrule

 \end{tabular}
    
\end{table}

\subsection{Learning Manifold Dimensions with CVAEs}

Turning to CVAEs, we provide empirical support for Theorem~\ref{compression_cvae} using synthetic datasets with varying $t$, the effective dimension of the conditioning variable $c$.  For this purpose, we set $c$ to be the first $t$ dimensions of $u$ by equating $G_c$ to a $t$-dim identity matrix $I_t$, and let $d = \kappa = 20$, and $r = 10$. Table~\ref{tab:cvae_varied_t} shows the results, whereby the CVAE correctly learns that $\mbox{AD} = r - t$ across all values of $t$.




Meanwhile, on MNIST and Fashion MNIST we train CVAEs with the class label of each image as the conditioning variable, and the results with $\kappa = 32$ are in Table~\ref{tab:cvae_mnist}. By comparing with Table~\ref{tab:vae_latent_mnist}, we observe that the CVAE model for MNIST has a lower AD and -ELBO, suggesting that when conditioned on labels, the model can more easily confine the resulting representations to a lower-dimensional manifold.  In contrast, for FashionMNIST the CVAE/VAE results are more similar, indicating that the class labels provide marginal benefit, possibly because their visual complexity and manifold structure is more similar across classes.   We also provide an example of the CVAE encoder variances produced on the MNIST dataset in Section~B of the appendix.



\begin{table}[]
    \centering
    \caption{CVAE latent compression showing $\text{AD} = r - t$ exactly in support of Theorem~\ref{compression_cvae}. Here $c$ is the first $t$ dimensions of $u$, i.e. $G_c=I_t$, $d = \kappa = 20$, and $r = 10$.}
    \label{tab:cvae_varied_t}
    \begin{tabular}{cccccc}
\toprule
    $t$ & -ELBO & Recon & $\mathbb{KL}$ & $\gamma$ & AD\\
    \midrule
    1 & -31.41 & 4.61$\times 10^{-2}$ & 33.26 & 2.4$\times 10^{-3}$ & 9\\
    3 & -36.67 & 4.66$\times 10^{-2}$ & 27.78 & 2.4$\times 10^{-3}$ & 7\\
    5 & -42.78 & 4.86$\times 10^{-2}$ & 20.81 & 2.6$\times 10^{-3}$ & 5\\
    7 & -52.39 & 4.29$\times 10^{-2}$ & 13.72 & 2.2$\times 10^{-3}$ & 3\\
    9 & -62.25 & 3.84$\times 10^{-2}$ & 6.07 & 2$\times 10^{-3}$ & 1\\
    \bottomrule
\end{tabular}
\end{table}

\begin{table}[]
\centering
\caption{CVAE latent compression in real datasets with $\kappa = 32$. AD is averaged over all the classes.}
\label{tab:cvae_mnist}
\begin{tabular}{cccccc}
\toprule
              & AD & Recon & $\mathbb{KL}$ & $\gamma$ & -ELBO      \\ \midrule
MNIST         & 12 & 6.044 & 81.672      & 0.0063   & -1489.42 \\
Fashion MNIST & 9  & 8.773 & 54.552      & 0.0102   & -1239.09 \\ \bottomrule
\end{tabular}
\end{table}

\subsection{Adaptive CVAE Active Dimensions within a Dataset} \label{sec:adaptive_active_dims}

Continuing with CVAE experimentation, we now turn to verifying aspects of Section~\ref{sec:adaptive_dim}, namely, the ability to adaptively learn active dimensions that vary regionally within a single dataset composed of a union of low-dimensional ground-truth manifolds.  To this end, we choose $c$ as a \textit{discrete} indicator of each source manifold, 5 in total with $r\in \{1,...,5\}$ latent dimensions within each respectively.  Table~\ref{tab:adaptive_discrete} shows the results using $d = 20$ and $\kappa = 40$, noting that both the AD and -ELBO values are class-conditional within a single dataset. Also, we report performance both with and without a special attention layer in the decoder that helps to selectively determine which dimensions should be active on a sample-by-sample basis.  These results indicate that, when equipped with a suitably flexible decoder network, the CVAE has the ability to adaptively learn AD values that correctly align with data lying on a union of manifolds.





\begin{table}[]
\centering
\caption{Adaptively learning CVAE active dimension involving data lying on a union of 5 manifolds with $r\in \{1,\ldots,5\}$, and $d = 20, \kappa = 40$.  A discrete $c$  labels each manifold/class, and the AD values and -ELBO are computed on a class-conditional basis.}
    \label{tab:adaptive_discrete}
\begin{tabular}{cccccc}
\toprule
$r$ & True & AD without & AD with & -ELBO   without          & -ELBO  with         \\
        &  AD &  attention   &   attention      &  attention & attention \\ \midrule
1                & 1                 & 4                & 1                & -102.25           & -114.22        \\
2                & 2                 & 4                & 2                & -62.42            & -99.81         \\
3                & 3                 & 4                & 3                & -60.13            & -74.28         \\
4                & 4                 & 4                & 4                & -28.06            & -50.36         \\
5                & 5                 & 4                & 5                & -7.58             & -59.25         \\ \bottomrule
\end{tabular}
    
\end{table}

To further explore adaptive active dimensions, we also consider the case where $c$ is a continuous variable with $t \in \{2,4,6,8,10 \}$ varying from sample-to-sample (this is achieved by using the first $\{2,4,6,8,10\}$ dimensions of $u$ as the conditioning variable).  To maintain a constant size of $c \in \mathbb{R}^{10}$ we apply zero-padding.  Using $r = 12$, $d = 20$, and $\kappa = 20$, results are shown in Table~\ref{tab:adaptive_continuous}, again with and without the attention layer that exploits $c$ to correctly learn the optimal AD across all values of $t$.


\begin{table}[]
\centering
\caption{Adaptively learning active dimensions involving a continuous $c$ and associated $t\in \{2,4,6,8,10\}$ varying within a single dataset; also $r = 12, d=20, \kappa=90$.  The AD values and -ELBO are computed on a class-conditional basis.}
    \label{tab:adaptive_continuous}
\begin{tabular}{cccccc}
\toprule
$t$ & True & AD without & AD with & -ELBO   without          & -ELBO  with         \\
        &  AD &  attention   &     attention    &  attention & attention \\ \midrule
2   & 10                & 10               & 10               & -9.69             & -41.49     \\
4   & 8                & 10                & 8                & -33.71            & -20.52         \\
6   & 6                & 10                & 6                & -27.10            & -73.26         \\
8   & 4                & 10                & 4                & -50.70            & -80.64         \\
10  & 2                & 10                & 2                & -61.77            & -55.14         \\ \bottomrule
\end{tabular}
    
\end{table}

\subsection{The Impact of $\gamma$ Initialization on Model Training}\label{subsec:init_gamma_exp}

We next investigate how the initialization of $\gamma$ impacts VAE/CVAE model convergence as related to the discussion in Section~\ref{subsec:gamma}. Results with synthetic data and $d=20$, $\kappa=20$, $r=10$, and for the CVAE, $t=5$ are shown in Table~\ref{tab:init_gamma_relu} as the initial $\gamma$ is varied.  Here we observe that for the VAE model, the correct number of active dimensions is only learned when $\gamma$ is initialized sufficiently small.  In contrast, for the CVAE we include results both with and without a parameterized prior. However, given an adequately-sized decoder, both models perform similarly as would be expected (note that we used the same decoder for both models, so the parameter-free prior model was technically not strictly equivalent per Remark \ref{rmk:equal_prior}).  And with the exception of the $\log \gamma = 20$ case when the parameterized CVAE prior model falls into a local minima, they are both able to correctly learn the number of active CVAE dimensions as $\mbox{AD} = r - t = 5$, consistent with Theorem \ref{compression_cvae}.  In comparison with the VAE, the CVAE models are likely less sensitive to the initial $\gamma$ because the number of active latent dimensions that need to be learned is half that of the VAE (5 versus 10).

\begin{table}[]
\centering
\caption{Learned active dimensions as initial $\gamma$ is varied, with $d = 20, r = 10, t = 5, \kappa=20$.}
\label{tab:init_gamma_relu}
\begin{tabular}{ccccccc}
\toprule
Init $\log \gamma$ & \multicolumn{2}{c}{VAE} & \multicolumn{2}{c}{CVAE $p(z)$} & \multicolumn{2}{c}{CVAE $p_\theta(z|c)$} \\ \midrule
                   & AD        & -ELBO       & AD            & -ELBO           & AD             & -ELBO            \\
                   \cmidrule(r){2-3} \cmidrule(r){4-5} \cmidrule(r){6-7}
-20                & 10        & -28.39       & 5              & -41.20           & 5              & -40.72            \\
-10                & 9        & -28.57         & 5           & -44.53          & 5              & -45.25            \\
0                  & 8        & -27.56       & 5             & -44.38          & 5              & -45.2            \\
10                 & 3         & -13.89       & 5            & -43.72          & 5              & -43.66            \\
20                 & 1         & -1.7         & 5          & -45.22           & 4              & -37.85            \\ \bottomrule
\end{tabular}

\end{table}

\subsection{Performance Degradation Using Shared Weights between Encoder and Prior}\label{subsec:share_weight_exp}

Finally, we provide empirical support for Theorem~\ref{thm:share_weight} relating to the potential impact of weight-sharing within sequential models. We generate sequences via an autoregressive-moving-average (ARMA) process, and we define the conditioning variable $c_l$ at time $l$ as the concatentation of the previous 5 timepoints within each sequence.  Results training a CVAE model to generate analogous sequences are shown in Table~\ref{tab:share_weight}.  We observe that when the encoder and prior share parameters (first row) the -ELBO cannot be significantly reduced, largely because the reconstruction error remains high.  In contrast, without shared weights (second row) the model successfully drives the reconstruction error and -ELBO to much smaller values as might be expected based on Theorem~\ref{thm:share_weight}. Of course in larger, more complex models there could be other mitigating factors, such as overfitting risks, that might allow shared weights to at times perform relatively better.

\begin{table}[]
    \centering
    \caption{Impact of weight sharing between encoder and prior on sequential data.}
    \label{tab:share_weight}
    \begin{tabular}{ccccc}
\toprule
     Shared Weights & -ELBO & Recon & $\mathbb{KL}$ & $\gamma$ \\
    \midrule
    True & -2.49 & 0.374 & 18.09 & 0.012 \\
    False & -45.015 & 1.81$\times 10^{-5}$ & 175.99  & 7.252$\times 10^{-7}$ \\
    \bottomrule
\end{tabular}
    
\end{table}


\subsection{Full vs.~Diagonal Encoder Covariances}

Our theoretical contributions from Section \ref{sec:manifold_learning} were predicated on properties of VAE/CVAE \textit{global} optima assuming sufficient model capacity to represent ground-truth manifolds.  However, we have spoken relatively little about explicit strategies for steering optimization trajectories away from bad \textit{local} solutions that may have less desirable characteristics.  In this regard, analysis from \cite{wipf2023} suggests that all else being equal, a \textit{full} (as opposed to diagonal) VAE encoder variance may provide a beneficial selective smoothing effect capable of guiding training iterations towards global solutions, mitigating the risk of converging to bad local alternatives.  

We now verify this theoretical prediction using MNIST data and the same experimental setup used to produce Table \ref{tab:vae_latent_mnist}.  The only modification is to accommodate the full covariance model, where we used the encoder to produce the Cholesky decomposition of a general covariance matrix.  Per this revised setup, results comparing equivalent-capacity VAEs (that only differ in the final encoder layer used to compute respective covariances, i.e., full vs.~diagonal) are displayed in Table \ref{tab:full_cov}.  

From these results we observe that all else being equal, the full covariance encoder model does in fact produce a significantly lower VAE loss. Moreover, the reconstruction error is lower even while using a fewer number of active latent dimensions. This closely aligns with predictions from \cite{wipf2023}, and complements our general findings here that VAE minima tend to align with optimal sparse representations, at least to the extent that good minima can be found.

\begin{table}[]
\centering
\caption{Comparing equivalent-capacity VAEs with diagonal vs.~full encoder covariances.  Setup is the same as Table \ref{tab:vae_latent_mnist} using MNIST data and $\kappa = 32$.}
\label{tab:full_cov}
\begin{tabular}{cccc}
\toprule
              & AD & Recon &  -ELBO      \\ \midrule
Full         & 8 & 4.5   & -1756 \\
Diagonal & 13  &  7.6   & -1205 \\ \bottomrule
\end{tabular}
\end{table}

\section{Conclusion}
In this paper we provide several insights into the behavior of VAEs and CVAEs applied to data lying on low-dimensional manifolds. This includes a formal demonstration that both VAE and CVAE global minima can learn manifold dimensions underlying the data, including those manifolds that have been modulated by conditioning variables.  We also explore common CVAE design choices that can have practical implications when applying to various specialized applications, such as sequential data or data composed of a union of manifolds.  That being said, beyond the attention mechanism we mentioned for learning region-specific active latent dimensions, we have not thoroughly explored what types of decoder inductive biases might best align with real-world manifold data.  This represents a reasonable direction for future work.

\bibliographystyle{plain}
\bibliography{ref}

\newpage
\appendix

\section*{Appendix}

\section{Details of Model Architectures and Implementation}

\paragraph{Synthetic Data} For the experiments with VAE models, both the encoder and decoder are defined as a Multi Layer Perceptron (MLP) with a single hidden layer. For all the experiments with CVAE models except Section 4.6, the encoder first processes conditioning variable $c$ via an MLP, then concatenates the output and samples $x$ as the input of another MLP. The decoder processes $c$ in the same way and then uses an MLP to decode the latent variable. In Section 4.4, the attention layer we use in the decoder is a trainable vector which is applied as the weight of the latent vector at the top layer. In Section 4.6, both the encoder and decoder are LSTMs with one hidden layer.

\paragraph{Real Data} Both the encoder and decoder use two ResNet blocks to process MNIST/ Fashion MNIST images. Each encoder block is a residual network which contains two $3\times 3$ Conv-BN-ReLU modules in its main branch and one $1\times 1$ Conv-BN module in its shortcut.
The decoder block contains a single-layer ConvNet residual block followed by a ConvTranspose layer.

\paragraph{Resources} We conduct our experiments on an Amazon Web Services g4dn.12xlarge EC2 instance, which provides 4 T4 GPUs. We estimate that the time to run through all experiments in this paper once would cost 20 GPU-hours. The research activity for this paper cost around 100 GPU-hours in total.

\section{Encoder Variance Illustration}
To show the active dimensions visually, here we report the encoder variances both on synthetic data and the MNIST dataset. Note that when the value of the encoder variance is less than $0.05$, we categorize the corresponding dimension as active for VAE models; for CVAEs we analogously require that the encoder/prior variance ratio is less than $0.05$.  Note however in Table 1 and 2 below, the number of active dimensions is quite obvious given the clear clustering of variance values.

\begin{table}[h!]
\centering
\caption{VAE encoder variance matrix on synthetic data associated with Table~1 of the main text, where $\kappa = 20, d=30, r=6$ and we find the number of active dimensions is 6. The estimated active dimensions are in blue.}
\begin{tabular}{ccccc}
\toprule
\textcolor{blue}{0.0080} & \textcolor{blue}{0.0018} & 1.0000 & 1.0000 & 1.0000 \\
\textcolor{blue}{0.0027} & \textcolor{blue}{0.0031} & 1.0000 & 1.0000 & 1.0000 \\
1.0000 & \textcolor{blue}{0.0087} & 1.0000 & \textcolor{blue}{0.0141} & 1.0000 \\
1.0000 & 1.0000 & 1.0000 & 1.0000 & 1.0000 \\ \bottomrule
\end{tabular}
\end{table}

\begin{table}[h!]
\centering
\caption{CVAE Encoder variance matrix of the CVAE model on MNIST dataset from Table~4 of the main text, where $\kappa = 32$ and  the number of active dimensions is 12. The estimated active dimensions are labeled in blue.}
\begin{tabular}{llll}
\toprule
\textcolor{blue}{3.6159e-03}  & 9.6320e-01  & \textcolor{blue}{7.6566e-04}  & \textcolor{blue}{3.5173e-04}  \\
9.8518e-01  & 9.6739e-01  & 9.6077e-01  & \textcolor{blue}{8.1020e-04}  \\
9.8065e-01  & 9.7336e-01  & \textcolor{blue}{3.7781e-03}  & \textcolor{blue}{7.1394e-04}  \\
9.6985e-01  & \textcolor{blue}{6.1294e-03}  & 9.7449e-01  & 9.8012e-01  \\
\textcolor{blue}{7.8233e-04}  & 9.7318e-01  & 9.8596e-01  & \textcolor{blue}{2.4359e-04}  \\
9.7785e-01  & 9.7737e-01  & 9.7315e-01  & 9.8431e-01  \\
9.2616e-01  & 9.8335e-01  & 9.6775e-01  & \textcolor{blue}{1.2756e-03}  \\
\textcolor{blue}{1.0324e-03}  & 9.6723e-01  & 9.6046e-01  & \textcolor{blue}{2.1289e-03} \\ \bottomrule
\end{tabular}
\end{table}

\section{Proof of Theorem 1}
\label{proof_th1}

\paragraph{Summary of the Proof} We define three categories based on the number of active dimensions and the rate of their encoder variance. Note that any possible VAE optimum has to fall into one of the following three categories: the number of dimensions whose encoder variance $\sigma_z^2(x, \phi_\gamma^\ast) = O(\gamma)$ is either greater than $r$, equal to $r$ or less than $r$. The proof's logic flow is:

\begin{enumerate}
    \item When the number of active dimensions whose encoder variance $\sigma_z^2(x) = O(\gamma)$ is less than $r$, the reconstruction error will increase at a rate of $O(\frac{1}{\gamma})$, thus the cost cannot reach the optimum. This is proven in Section~\ref{ch:ext_of_active};
    \item When the stated dimension number equals $r$, the optimal cost is exactly $(d-r)\log \gamma + O(1)$. The corresponding proof is in Section~\ref{sec:ad_eq_r};
    \item When the stated dimension number is greater than $r$, denoted as $m > r$, the cost is $(d-m)\log \gamma + O(1) > (d-r)\log \gamma + O(1)$ as shown in Section~\ref{sec:ad_g_r}.
    \end{enumerate}

The $O(\gamma)$ rate of the reconstruction error also follows naturally from these results.

\subsection{The number of active dimensions whose encoder variance $\sigma^2_z(x) = O(\gamma)$ is less than $r$}
\label{ch:ext_of_active}

The main idea is to link the gap between a large $\sigma_z$ and large reconstruction error. For a given $z_0$, $\mu_x(z_0)$ will equal some $x_0$ such that $||x_0 - \mu_x(z_0)||^2 = 0$. But for other choices from $\mathcal{X}$ where $x \neq x_0$, we have $||x - \mu_x(z_0)||^2 > 0$ leading to the positive expectation term $\int_z q(z|x) ||x - \mu_x(z)||^2 dz$. To minimize such positive error, we need to lower the density $q(z|x)$ where $x \neq x_0$, which is a function of $\sigma_z$.

Suppose that the number of active dimensions whose encoder variance $\sigma^2_z(x) = O(\gamma)$ is less than $r$. In this section, we will show that under this assumption the model can't reach its global optimum, i.e. $\mathcal{L} \nrightarrow -\infty$. Remind that the cost of VAE is $$\mathcal{L}(\theta, \phi) = \int_\mathcal{X} \{-\mathbb{E}_{q_\phi (z|x)}[\log p_\theta(x|z)] + \mathbb{KL}[q_\phi(z|x)||p(z)]\} \omega_{gt}(dx)$$

We have

\begin{equation}
    \begin{aligned}
        2\mathcal{L}(\theta, \phi) &= \int_\mathcal{X} \{-2\mathbb{E}_{q_\phi (z|x)}[\log p_\theta(x|z)] + 2\mathbb{KL}[q_\phi(z|x)||p(z)]\} \omega_{gt}(dx) \\
        &= d \log (2\pi \gamma) + \int_\mathcal{X} \left\{\gamma^{-1} \mathbb{E}_{q_{\phi}(z|x)} [||x - \mu_x(z)||^2] + 2\mathbb{KL}(q_\phi(z|x)||p(z))\right\} \omega_{gt}(dx) \\
        &= d \log (2\pi \gamma) + \gamma^{-1}\int_\mathcal{X} \int_z q_{\phi}(z|x) [||x - \mu_x(z)||^2] dz\ \omega_{gt}(dx) \\
        & \quad + \int_\mathcal{X} 2\mathbb{KL}(q_\phi(z|x)||p(z)) \omega_{gt}(dx)
    \end{aligned}
\end{equation}

Following the two facts:

\begin{enumerate}
    \item Lebesgue measure on the real numbers is $\sigma$-finite.
    \item $z \in \mathbb{R}^\kappa$ and $x \in \mathcal{X}$, where $\mathcal{X}$ is a $r$-dimensional manifold embedded in $\mathbb{R}^d$.
\end{enumerate}

and referring to Fubini's theorem, we can switch the integration order of $\omega_{gt}(dx)$ and $dz$. Assume the components of $z \in \mathcal{Z}^\kappa \subseteq \mathbb{R}^\kappa$ is permutable. For a $r$-dimensional manifold, we can always use the first $r$ dimensions of $z$ to get $\varphi(x)$, i.e. once given r-dimensional information, there always exists a decoder, s.t. $\mu_x(z_{1:r}) = \mu_x(z)$. Denote by $\mu_z(x)_{1:r}$ and $\sigma_z^2(x)_{1:r}$ the mean and covariance matrix of the first $r$ dimension of $z$. After switching the integration order, we have

\begin{equation}\label{eq:expand_density}
\begin{aligned}
    & \int_\mathcal{X} \frac{1}{\gamma} \int_z q_{\phi}(z|x)  [||x - \mu_x(z)||^2] dz\ \omega_{gt}(dx) +  \int_\mathcal{X} \left[ d \log (2\pi \gamma) + 2\mathbb{KL}(q_\phi(z|x)||p(z))\right] \omega_{gt}(dx)\\
    = & \frac{1}{\gamma} \int_{z} \int_\mathcal{X} q_{\phi}(z|x)  [||x - \mu_x(z_{1:r})||^2] \omega_{gt}(dx)dz + \int_\mathcal{X} \left[d \log \gamma + 2\mathbb{KL}(q_\phi(z|x)||p(z)) + O(1)\right] \omega_{gt}(dx)\\
    = & \frac{1}{\gamma} \int_{z \in \mathcal{Z}^r} \int_\mathcal{X} \frac{1}{\sqrt{(2\pi)^{r} |\sigma_z^2(x)_{1:r}|}}e^{-\frac{1}{2}(z - \mu_z(x)_{1:r})^T\sigma_z^{-2}(x)_{1:r}(z - \mu_z(x)_{1:r})}  [||x - \mu_x(z)||^2] \omega_{gt}(dx)dz +\\
        & \int_\mathcal{X} \left[d \log \gamma + 2\mathbb{KL}(q_\phi(z|x)||p(z)) + O(1)\right] \omega_{gt}(dx)\\
\end{aligned}
\end{equation}

\subsubsection{Analyze the density with respect to $\sigma_z(x)_{1:r}$ and $z_{1:r} - \mu_z(x)_{1:r}$}

Next, for the integral over $\mathcal{X}$ in the first term in (\ref{eq:expand_density}), we examine a certain $z_{1:r} \in \mathcal{Z}^r$ and view it as a constant. Since $\mu_x$ is a deterministic function, $\mu_x(z_{1:r})$ is also constant. The log-density on $z_{1:r}$ is

$$
\frac{r}{2}\log (\frac{1}{2\pi}) + \frac{1}{2} \log \frac{1}{|\sigma^2_z|} - \frac{1}{2}(z_{1:r} - \mu_z(x)_{1:r})^T\sigma_z^{-2}(z_{1:r} - \mu_z(x)_{1:r})
$$

Take the derivative of $\sigma_z^2$, we have

$$
- \frac{ \sigma_z^{-2}}{2} + \frac{1}{2} \sigma_z^{-2} (z_{1:r} - \mu_z(x)_{1:r})(z_{1:r} - \mu_z(x)_{1:r})^T \sigma_z^{-2}
$$

When $\sigma^2_z$ is smaller than $(z_{1:r} - \mu_z(x)_{1:r})(z_{1:r} - \mu_z(x)_{1:r})^T$, the second term's rate is larger. Thus the density is monotonically increasing when $\sigma^2_z \prec (z_{1:r} - \mu_z(x)_{1:r})(z_{1:r} - \mu_z(x)_{1:r})^T$ and monotonically decreasing when $\sigma^2_z \succ (z_{1:r} - \mu_z(x)_{1:r})(z_{1:r} - \mu_z(x)_{1:r})^T$. Note that $\mu_x$ is $L$-Lipschitz continuous, so we have $L|z_{1:r} - \mu_z(x)_{1:r}| \geq |\mu_x(z_{1:r}) - \mu_x(\mu_z(x)_{1:r})| = |\mu_x(z_{1:r}) - x|$. The equality comes from the fact that we can choose optimal $\mu_z$ and $\mu_x$, s.t. $\mu_x(\mu_z(x)_{1:r}) = x$. 

Now we can divide $x \in \mathcal{X}$ into four cases and we assume all the four disjoint cases exist when analyzing, otherwise the integration over corresponding domain is 0 which would not affect our result. The four cases are as follows:

\begin{enumerate}
    \item $\mathcal{X}_1(z_{1:r}) = \{ x: \sigma_z^2(x)_{1:r} \prec (z_{1:r} - \mu_z(x)_{1:r})(z_{1:r} - \mu_z(x)_{1:r})^T\} \cap \{x: ||z_{1:r} - \mu_z(x)_{1:r}|| = +\infty$\}
    \item $\mathcal{X}_2(z_{1:r}) = \{ x: \sigma_z^2(x)_{1:r} \prec (z_{1:r} - \mu_z(x)_{1:r})(z_{1:r} - \mu_z(x)_{1:r})^T\} \cap \{x: ||z_{1:r} - \mu_z(x)_{1:r}|| < +\infty$\}
    \item $\mathcal{X}_3(z_{1:r}) = \{ x: (z_{1:r} - \mu_z(x)_{1:r})(z_{1:r} - \mu_z(x)_{1:r})^T \preceq  \sigma_z^2(x)_{1:r} < \infty \}$
    \item $\mathcal{X}_4(z_{1:r}) = \{ x: (z_{1:r} - \mu_z(x)_{1:r})(z_{1:r} - \mu_z(x)_{1:r})^T \preceq  \sigma_z^2(x)_{1:r} = \infty \}$
\end{enumerate}

We have $\mathcal{X}_1(z_{1:r}) \cup \mathcal{X}_2(z_{1:r}) = \{ x: \sigma_z^2(x)_{1:r} \prec (z_{1:r} - \mu_z(x)_{1:r})(z_{1:r} - \mu_z(x)_{1:r})^T\}$ and $\mathcal{X}_3(z_{1:r}) \cup \mathcal{X}_4(z_{1:r}) = \{ x: \sigma_z^2(x)_{1:r} \succeq (z_{1:r} - \mu_z(x)_{1:r})(z_{1:r} - \mu_z(x)_{1:r})^T\}$. Thus $\mathcal{X}_1(z_{1:r}) \cup \mathcal{X}_2(z_{1:r}) \cup \mathcal{X}_3(z_{1:r}) \cup \mathcal{X}_4(z_{1:r})$ cover the whole space of $x$ related to $z_{1:r}$, i.e. $\mathcal{X}(z_{1:r})$.

\textit{(i) $\mathcal{X}_1(z_{1:r}) = \{ x: \sigma_z^2(x)_{1:r} \prec (z_{1:r} - \mu_z(x)_{1:r})(z_{1:r} - \mu_z(x)_{1:r})^T\} \cap \{x: ||z_{1:r} - \mu_z(x)_{1:r}|| = +\infty$\} }

Denote $\sigma^2_l$ as the lower bound of $\sigma^2_z$'s eigenvalues, which cannot approach 0 by our assumption. $\sigma_z < +\infty$. The integral over $\mathcal{X}_1(z_{1:r})$

$$
\begin{aligned}
    & \int_{\mathcal{X}_1(z_{1:r})} \frac{1}{\sqrt{|\sigma^2_z|}}e^{-\frac{1}{2}(z_{1:r} - \mu_z(x)_{1:r})^T\sigma_z^{-2}(z_{1:r} - \mu_z(x)_{1:r})}  [||x - \mu_x(z_{1:r})||^2] \omega_{gt}(dx)_r \\
    \leq & \int_{\mathcal{X}_1(z_{1:r})} \frac{1}{\sigma_l^r}e^{-\frac{1}{2}(z_{1:r} - \mu_z(x)_{1:r})^T\sigma_z^{-2}(z_{1:r} - \mu_z(x)_{1:r})}  [L^2||z_{1:r} - \mu_z(x)_{1:r})||^2] \omega_{gt}(dx)_r
\end{aligned}
$$

will approach 0 as $ ||z_{1:r} - \mu_z(x)_{1:r}|| = +\infty$. Thus $\int_{z_{1:r}}0dz=0$

\textit{(ii) $\mathcal{X}_2(z_{1:r}) = \{ x: \sigma_z^2(x)_{1:r} \prec (z_{1:r} - \mu_z(x)_{1:r})(z_{1:r} - \mu_z(x)_{1:r})^T\} \cap \{x: ||z_{1:r} - \mu_z(x)_{1:r}|| < +\infty$\}}

Denote $N = \max_x\{||z_{1:r} - \mu_z(x)_{1:r}||^2\}$ and $\mathcal{X}_2^\alpha(z_{1:r}) = \{x: ||x - \mu_x(z_{1:r})||^2 > \alpha\} \cap \mathcal{X}_2(z_{1:r})$, where $\alpha \nrightarrow 0$. If for any $\alpha$, $\mathcal{X}_2^\alpha(z_{1:r}) = \emptyset$, we have for all $x \in \mathcal{X}_2(z_{1:r})$, $x = \mu_x(z_{1:r})$. However, if $\mu_x(x)_r \in \mathcal{X}_2(z_{1:r})$, i.e. $\mu_x(z_{1:r})$ satisfies $\sigma^2_z(\mu_x(z_{1:r})) \prec (z_{1:r} - \mu_z(\mu_x(z_{1:r})))(z_{1:r} - \mu_z(\mu_x(z_{1:r})))^T$. We can find a pair of $\mu_x$ and $\mu_z$, e.g. identity mapping, s.t. $\mu_z(\mu_x(z_{1:r})) = z_{1:r}$ and $\sigma^2_z(\mu_x(z_{1:r})) < 0$, which is impossible. Thus $\mu_x(z_{1:r}) \notin \mathcal{X}_2(z_{1:r})$ and $\mathcal{X}_2(z_{1:r}) = \emptyset$. Thus, once $\mathcal{X}_2(z_{1:r}) \neq \emptyset$, there exists an $\alpha$, s.t. $\mathcal{X}_2^\alpha(z_{1:r}) \neq \emptyset$. The integral over $\mathcal{X}_2(z_{1:r})$

$$
\begin{aligned}
    & \int_{\mathcal{X}_2(z_{1:r})} \frac{1}{\sqrt{|\sigma^2_z|}}e^{-\frac{1}{2}(z_{1:r} - \mu_z(x)_{1:r})^T\sigma_z^{-2}(z_{1:r} - \mu_z(x)_{1:r})}  [||x - \mu_x(z_{1:r})||^2] \omega_{gt}(dx)_r \\
    \geq & \int_{\mathcal{X}_2(z_{1:r})} \frac{1}{\sigma_l^r}e^{-\frac{1}{2}\sigma_l^{-2r}N}  [||x - \mu_x(z_{1:r})||^2] \omega_{gt}(dx)_r\\
    = &  \frac{1}{\sigma_l^{r}}e^{-\frac{1}{2}\sigma_l^{-2r}N} [\int_{\mathcal{X}_2^\alpha(z_{1:r})}  [||x - \mu_x(z_{1:r})||^2] \omega_{gt}(dx)_r + \int_{ (\mathcal{X}_2^\alpha(z_{1:r}))^c}  [||x - \mu_x(z_{1:r})||^2] \omega_{gt}(dx)_r] \\
    \geq & \frac{\alpha}{\sigma_l^{r}}e^{-\frac{1}{2}\sigma_l^{-2r}N}
\end{aligned}
$$

The last inequality comes from the fact that $\varpi(\mathcal{X}_2^\alpha(z_{1:r})) \geq 1$ where $\varpi$ is a counting measure and the non-negativity of the second term.

\textit{(iii) $\mathcal{X}_3(z_{1:r}) = \{ x: (z_{1:r} - \mu_z(x)_{1:r})(z_{1:r} - \mu_z(x)_{1:r})^T \preceq  \sigma_z^2(x)_{1:r} < \infty \}$ }

In this case the density is monotonically decreasing with $\sigma_z$. Since $\sigma_z \neq +\infty$, denote $\sigma^2_u$ as the upper bound of the eigenvalues of $\sigma_z^2$. It can also bound $||z_{1:r} - \mu_z(x)_{1:r}||^2$. Use the same strategy in $(ii)$, define $\mathcal{X}_3^{\alpha'}(z_{1:r}) = \{x: ||x - \mu_x(z_{1:r})||^2 > \alpha'\} \cap \mathcal{X}_3(z_{1:r})$. If $\mathcal{X}_3(z_{1:r}) \neq \emptyset$, we have

$$
\begin{aligned}
    & \int_{\mathcal{X}_3(z_{1:r})} \frac{1}{\sqrt{|\sigma^2_z|}}e^{-\frac{1}{2}(z_{1:r} - \mu_z(x)_{1:r})^T\sigma_z^{-2}(z_{1:r} - \mu_z(x)_{1:r})}  [||x - \mu_x(z_{1:r})||^2] \omega_{gt}(dx)_r \\
    \geq & \int_{\mathcal{X}_3(z_{1:r})} \frac{1}{\sigma_u^{r}}e^{-\frac{r}{2}\sigma_u^{-2r+2}}  [||x - \mu_x(z_{1:r})||^2] \omega_{gt}(dx)_r \\
    = &  \frac{1}{\sigma_u^{r}}e^{-\frac{r}{2}\sigma_u^{-2r+2}} [\int_{\mathcal{X}_3^{\alpha'}(z_{1:r})}  [||x - \mu_x(z_{1:r})||^2] \omega_{gt}(dx)_r + \int_{ (\mathcal{X}_3^{\alpha'}(z_{1:r}))^c}  [||x - \mu_x(z_{1:r})||^2] \omega_{gt}(dx)_r] \\
    \geq & \frac{\alpha'}{\sigma_u^{r}}e^{-\frac{r}{2}\sigma_u^{-2r+2}} \\
\end{aligned}
$$

\textit{(iv) $\mathcal{X}_4(z_{1:r}) = \{ x: (z_{1:r} - \mu_z(x)_{1:r})(z_{1:r} - \mu_z(x)_{1:r})^T \preceq  \sigma_z^2(x)_{1:r} = \infty \}$ }

In this case the density is monotonically decreasing with $\sigma_z$, and the dominant factor is $\frac{1}{\sqrt{|\sigma_z^2(x)_{1:r}|}}$. Since $\sigma_z$ is arbitrarily large, it is obvious that $\sqrt{|\sigma_z^2|} > tr(\sigma_z^2) \geq ||z_{1:r} - \mu_z(x)_{1:r}||^2 \geq \frac{1}{L^2} ||x - \mu_x(z_{1:r})||^2$. Note that $|\cdot| = det(\cdot)$.

The integral over $\mathcal{X}_4(z_{1:r})$

$$
\begin{aligned}
    & \int_{\mathcal{X}_4(z_{1:r})} \frac{1}{\sqrt{|\sigma^2_z|}}e^{-\frac{1}{2}(z_{1:r} - \mu_z(x)_{1:r})^T\sigma_z^{-2}(z_{1:r} - \mu_z(x)_{1:r})}  [||x - \mu_x(z_{1:r})||^2] \omega_{gt}(dx)_r \\
    \leq & \int_{\mathcal{X}_4(z_{1:r})} \frac{L^2||z_{1:r} - \mu_z(x)_{1:r})||^2}{\sqrt{|\sigma_z^2(x)_{1:r}|}}  \omega_{gt}(dx)_r\\
\end{aligned}
$$

will approach $0$ as $\sigma_z^2 \rightarrow \infty$.

\subsubsection{Analyze the existence of the above cases and get a lower bound}

We have $\mathcal{X} = \mathcal{X}_1 \cup \mathcal{X}_2 \cup \mathcal{X}_3 \cup \mathcal{X}_4$, where $\mathcal{X}_i = \cup_{z_{1:r}} \mathcal{X}_i(z_{1:r}), i= 1, 2, 3, 4$. In $\mathcal{X}_1 \cup \mathcal{X}_4$, the integral is 0. To get a lower bound of (\ref{eq:expand_density}), we need to prove $\mathcal{X}_2 \cup \mathcal{X}_3 \neq \emptyset$, i.e. there must exists $z_{1:r}$ such that $x \in \{\sigma_z^2(x)_{1:r} < \infty\} \cap \{||z_{1:r} - \mu_z(x)_{1:r}|| < \infty\}$ exists. 

For $\sigma_z(x)_r$, if $\sigma_z(x)_r = \infty$, then in the KL term the trace $tr(\sigma_z^2(x)_{1:r}) = +\infty$ which cannot be offset by $-\log |\sigma_z^2(x)_{1:r}|$. Thus to minimize loss, $\sigma_z < \infty$.

For $||z_{1:r} - \mu_z(x)_{1:r}|| < \infty$, with $L$-Lipschitz continuity, for any $z_{1:r}^*$, we can find a $x^* \in \mathcal{X}$, s.t. $||z_{1:r}^* - \mu_z(x^*)_{1:r}|| = 0$. Denote $U_\delta(x)_r$ as a neighborhood of $x$ with the radius of $\delta$. For any $x \in U_\delta(x^*)$ , we have 

$$||\mu_z(x)_{1:r} - z_{1:r}^*|| = ||\mu_z(x)_{1:r} - \mu_z(x^*)_{1:r}|| \leq L||x - x^*|| \leq L\delta$$ 

So $U_\delta(x^*) \subset \mathcal{X}_2 \cup \mathcal{X}_3$. To get a positive lower bound, we need to prove there exists $x'$ and $\delta$, s.t. the image of $\mu_z(x')$ for $x \in  U_\delta(x')$ is with positive measure. If for any $x \in  U_\delta(x^1)$, $\mu_z(x)_{1:r} = z_{1:r}^1$, and for any $x \in  U_\delta(x^2)$, $\mu_z(x)_{1:r} = z_{1:r}^2$, which satisfy $\delta < ||x^2 - x^1|| \leq \frac{3}{2}\delta$, and $z_{1:r}^1 \neq z_{1:r}^2$. Note that can always choose a larger $\delta$ to get such pair of $\{x^1, x^2\}$. Then there exists $x^3 \in U_\delta(x^1) \cap U_\delta(x^2)$, $\mu_z(x^3)$ should equals $z_{1:r}^1$
 and $z_{1:r}^2$ simultaneously which is impossible. Thus, there must exists $x^*$, s.t. $\mu_z(U_\delta(x^*))$ has a positive measure. 

With the existence of $x^*$, such that $U_\delta(x^*) \subset \mathcal{X}_2 \cup \mathcal{X}_3$ and positive measured $\mu_z(U_\delta(x^*))$, we have

\begin{equation}
    \begin{aligned}
        & \frac{1}{\gamma} \int_{z_{1:r}} \int_\mathcal{X} \frac{1}{\sqrt{(2\pi)^{r} |\sigma^2_z|}}e^{-\frac{1}{2}(z_{1:r} - \mu_z(x)_{1:r})^T\sigma_z^{-2}(z_{1:r} - \mu_z(x)_{1:r})}  [||x - \mu_x(z_{1:r})||^2] \omega_{gt}(dx)_rdz_{1:r} +\\
        & \int_\mathcal{X} \left[ d \log \gamma + 2\mathbb{KL}(q_\phi(z|x)||p(z)) + O(1) \right] \omega_{gt}(dx)\\
        = & \frac{1}{\gamma} \int_{z_{1:r}} \int_{\mathcal{X}_2 \cup \mathcal{X}_3} \frac{1}{\sqrt{(2\pi)^{r} |\sigma^2_z|}}e^{-\frac{1}{2}(z_{1:r} - \mu_z(x)_{1:r})^T\sigma_z^{-2}(z_{1:r} - \mu_z(x)_{1:r})}  [||x - \mu_x(z_{1:r})||^2] \omega_{gt}(dx)_rdz_{1:r} +\\
        & \int_\mathcal{X} \left[ d \log \gamma + 2\mathbb{KL}(q_\phi(z|x)||p(z)) + O(1) \right] \omega_{gt}(dx)\\
        \geq & \frac{1}{\gamma} \int_{z_{1:r}} \min\{\frac{\alpha}{\sigma_l^{r}}e^{-\frac{1}{2}\sigma_l^{-2r}N} , \frac{\alpha'}{\sigma_u^{r}}e^{-\frac{r}{2}\sigma_u^{-2r+2}}\}  dz_{1:r} +\\
        & \int_\mathcal{X} \left[ d \log \gamma + 2\mathbb{KL}(q_\phi(z|x)||p(z)) + O(1) \right] \omega_{gt}(dx)\\
        = & \frac{C}{\gamma} + \int_\mathcal{X} \left[ d \log \gamma - \log |\sigma^2_z(x)| + O(1) \right] \omega_{gt}(dx)\\
    \end{aligned}
\end{equation}

Here denote $C = \int_{z_{1:r}} \min\{\frac{\alpha}{\sigma_l^{r}}e^{-\frac{1}{2}\sigma_l^{-2r}N} , \frac{\alpha'}{\sigma_u^{r}}e^{-\frac{r}{2}\sigma_u^{-2r+2}}\} dz_{1:r}$ for simplicity.

\subsubsection{Analyze the rate of the lower bound}

The first term $\frac{C}{\gamma}$ grows at a rate of O($\frac{1}{\gamma}$). Because $\sigma_z^2$ is at a lower rate than $\gamma$, we have $$O(d \log \gamma - \log |\sigma^2_z(x)|) < O(- (d - \kappa) \log \frac{1}{\gamma})$$ and the right part decreases at a rate of $\log \frac{1}{\gamma}$. When $\gamma \rightarrow 0$, $O(\frac{1}{\gamma}) > O(\log \frac{1}{\gamma})$, which means the increase from reconstruction term cannot be offset by the decrease from the KL term. Moreover, from the fact that $O(\frac{1}{\gamma}) > O(\log \frac{1}{\gamma})$, when $\gamma$ is small enough, the loss is 
monotonically decreasing with $\gamma$. 

Therefore, when $\gamma \rightarrow 0$, the lower bound cannot approach $-\infty$, which means at this case, the model can never achieve optimum. Thus, there must exist some active dimensions whose variance $\sigma^2_z(x)$ satisfies $\sigma^2_z(x) = O(\gamma)$ as $\gamma \rightarrow 0$ to reach the global optimum. We can learn from the expression of $C$ that as long as the number of such active dimensions whose encoder variance $\sigma^2_z(x) = O(\gamma)$ exceeds $r$, as $\gamma$ approaches zero, the reconstruction term is at most at the rate of $O(1)$. Next, we will show that when there exist at least $r$ such active dimensions, the VAE model's optimum is achievable.

\subsection{The number of active dimensions whose encoder variance $\sigma^2_z(x) = O(\gamma)$ equals $r$}
\label{sec:ad_eq_r}

In this section, we get an upper bound and a lower bound and show that both case the cost is $(d-r) \log \gamma + O(1)$. 

\subsubsection{An Upper Bound of ELBO}

\paragraph{Get an upper bound by Lipschitz}

We can write $z = \mu_z(x) + \varepsilon * \sigma_z(x)$, where $\varepsilon \sim N(0, I)$. Since decoder mean function $\mu_x(z; \theta)$ is $L$-Lipschitz continuous, we have:

\begin{equation}
    \begin{aligned}
    \label{eq:lip_ineq}
        & \mathbb{E}_{z \sim q_{\phi_{\gamma}}(z|x)}[||x - \mu_x(z)||^2] \\
         =& \mathbb{E}_{\varepsilon \sim N(0, I)} [||\mu_x(\mu_z(x)_{1:r}) - \mu_x(z_{1:r})||^2] \\
        \leq & \mathbb{E}_{\varepsilon \sim N(0, I)} [||L(\mu_z(x)_{1:r} - \mu_z(x)_{1:r} - \sigma_z(x)_{1:r} \varepsilon)||^2] \\
        = & \mathbb{E}_{\varepsilon \sim N(0, I)} [||L \sigma_z(x)_{1:r} \varepsilon||^2]\\
    \end{aligned}
\end{equation}

where the first equality comes from the fact that we can choose optimal encoder-decoder pairs such that $\mu_x(\mu_z(x)_{1:r}) = x$. Take it into $\mathcal{L}$,

\begin{equation}
\begin{aligned}
            &2\mathcal{L}(\sigma_z(x)_{1:r}, \gamma)\\
            =& \int_\mathcal{X} \left[ \mathbb{E}_{z \sim q_\phi(z|x)}[\frac{1}{\gamma} ||x - \mu_x(z)||_2^2] + d\log 2\pi \gamma - \log |\sigma^2_z(x)_{1:r}| - \log |\sigma^2_z(x)_{r+1:\kappa}| + O(1) \right] \omega_{gt}(dx)\\
            \leq& \frac{L^2}{\gamma} \int_\mathcal{X} \left[ \mathbb{E}_{\varepsilon \sim N(0, I)} [||\sigma_z(x)_{1:r} \varepsilon||^2] + d\log 2\pi \gamma - \log |\sigma^2_z(x)_{1:r}| - \log |\sigma^2_z(x)_{r+1:\kappa}| + O(1)\right] \omega_{gt}(dx)
\end{aligned}
\end{equation}

We get an upper bound of $\mathcal{L}$, denoted as $\mathcal{\tilde{L}}$.

\paragraph{Analysis of the Upper Bound $\mathcal{\tilde{L}}$}
\label{upper_bound}

Now we only pay attention to the upper bound $\mathcal{\tilde{L}}$ and try to prove that it is at a rate of $O((d-r)\log \gamma)$. 

We can get implicit optimal values of $\mathcal{\tilde{L}}$: $\gamma^*$ and $\sigma^*(x)_{1:r}^2$ by taking the derivative of $\mathcal{\tilde{L}}$ separately. 

We have optimal $\gamma^*$

\begin{equation}
\label{eq:opt_gamma}
    \gamma^* = \arg \min_\gamma \mathcal{\tilde{L}(\theta, \phi)} = \frac{L^2}{d} \mathbb{E}_{\varepsilon \sim N(0, I)} [||\sigma_z(x)_{1:r} \varepsilon||^2]
\end{equation}

and

$$
    \begin{aligned}
            \frac{\partial 2\mathcal{\tilde{L}}(\sigma_z(x)_{1:r}, \gamma)}{\partial \sigma_z(x)_{1:r}} &=  \frac{2L^2\sigma_z(x)_{1:r}}{\gamma} \mathbb{E}_{\varepsilon \sim N(0, I)}[\varepsilon\varepsilon^T] - 2\sigma_z(x)_{1:r}^{-1}\\
            & = \frac{2L^2\sigma_z(x)_{1:r}}{\gamma} - 2\sigma_z(x)_{1:r}^{-1} = 0
    \end{aligned}
$$
we have the optimal variance of $\mathcal{\tilde{L}}$:

\begin{equation}
    \label{gns}
    \sigma_z^*(x)_{1:r}^2 = \gamma \frac{I}{L^2}
\end{equation}

It shows that $\frac{1}{\sqrt{\gamma}} \sigma_z^*(x)_{1:r}  = O(1)$ when it reaches the optimal value. 

Take the optimal values into $\mathcal{\tilde{L}}$, then we get $\mathcal{\tilde{L}}$ as a function of $\gamma^*$ and $\sigma^*_z(x)_{1:r}$:

\begin{equation}
\label{eq:loss_ineq}
    \begin{aligned}
    & 2\mathcal{\tilde{L}}(\gamma^*, \sigma^*_z(x)_{1:r}) \\
    = & \int_\mathcal{X} \left[  \frac{L^2}{\gamma^*} \mathbb{E}_{\varepsilon \sim N(0, I)} [||\sigma_z(x)_{1:r} \varepsilon||^2] + d\log 2\pi \gamma^* - \log |\sigma^*_z(x)^2_{1:r}| - \log |\sigma_z(x)^2_{r+1:\kappa}| + O(1) \right] \omega_{gt}(dx)\\
    = &\int_\mathcal{X} \left[ d + d \log(2\pi \gamma^*) - \log |\gamma^* \frac{I}{L^2}| - \log |\sigma_z(x)^2_{r+1:\kappa}| + O(1) \right] \omega_{gt}(dx) \\
    = & d \log (2\pi \gamma^*) - r \log \gamma^*  - \log |\sigma_z(x)^2_{r+1:\kappa}| + O(1)
    \end{aligned}
\end{equation}

Define $\{\lambda_i\}_{i=1}^\kappa$ as the eigenvalues of $\sigma_z(x)$. Denote $\tilde{r}$ as the number of $\{\lambda_i\}_{i=r+1}^\kappa$ that will go to 0 as $\gamma^* \rightarrow 0$. We have 

\begin{equation}
\label{loss}
    \begin{aligned}
    & 2\mathcal{\tilde{L}}(\gamma^*)
    = d \log \gamma^* - r\log \gamma^* - 2\sum_{i=r+1}^{r+\tilde{r}} \log \lambda_i - 2\sum_{i=\tilde{r}+r+1}^\kappa \log \lambda_i + O(1) \\
    \end{aligned}
\end{equation}

To minimize (\ref{loss}), we want $\tilde{r}$ to be as small as possible and at the best equals 0 which is achievable. Since the rest $\kappa - r - \tilde{r} = \kappa - r$ dimensions are irrelevant to $\gamma$, at least will not approach 0 when $\gamma \rightarrow 0$, we can view them as constants. We have the loss equals

\begin{equation}
\label{eq:ub}
    (d-r) \log \gamma + O(1)
\end{equation}

\subsubsection{A Lower Bound of ELBO}

From \ref{ch:ext_of_active} we have analyzed the loss performance when there are less than $r$ active dimensions whose variance goes to zero at a rate no lower than $\gamma$. In this part, we focus on the case that $r$ latent dimensions are such active dimensions whose encoder variance goes to zero at a rate of $O(\gamma)$. Without loss of generality, we assume the first $r$ latent dimensions satisfy $\sigma^2_z(x)_{1:r} = O(\gamma)$ as $\gamma \rightarrow 0$. We have 

\begin{equation}
    \begin{aligned}
        2\mathcal{L}(\theta, \phi) = &\int_\mathcal{X} \{-2\mathbb{E}_{q_\phi (z|x)}[\log p_\theta(x|z)] + 2\mathbb{KL}[q_\phi(z|x)||p(z)]\} \omega_{gt}(dx) \\
        = & \int_\mathcal{X} \left\{\frac{1}{\gamma} \mathbb{E}_{q_{\phi}(z|x)} [||x - \mu_x(z)||^2] + d \log (2\pi \gamma) + 2\mathbb{KL}(q_\phi(z|x)||p(z)) \right\} \omega_{gt}(dx) \\
        \geq & \int_\mathcal{X} \left\{ d \log (2\pi \gamma) - \log |\sigma_z^2(x)| + O(1) \right\} \omega_{gt}(dx)\\
        = & \int_\mathcal{X} \left\{ d \log \gamma - \log |\sigma_z^2(x)_{1:r}| - \log |\sigma_z^2(x)_{r+1:\kappa}| + O(1) \right\} \omega_{gt}(dx) \\
        \geq & \int_\mathcal{X} \left\{ (d - r) \log \gamma - \log |\sigma_z^2(x)_{r+1:\kappa}| + O(1) \right\} \omega_{gt}(dx)\\
         = & (d - r) \log \gamma + O(1) \\
    \end{aligned}
\end{equation}

The inequalities come from the fact that the norm term is non-negative and the active dimensions' rate is no less than $\gamma$. For the last equality, we can use the strategy in (\ref{loss}). To minimize the lower bound, there should not be any active dimensions in these $r+1: \kappa$ dimensions. 

We get an upper bound and a lower bound at the same rate, i.e. $\log \gamma$ with $r$ active latent dimensions. Therefore, the original loss is also with $r$ active dimensions. We have the optimal cost for each $x$ equals

\begin{equation}
    (d - r) \log \gamma + O(1)
\end{equation}

So far, we have get the conclusion in Theorem~1 about the form of ELBO when $\gamma \rightarrow 0$, as well as the number and rate of active dimensions. Next, we show that the number of active dimensions can't be greater than $r$.

\subsection{When the number of active dimensions is greater than $r$}
\label{sec:ad_g_r}

Denote now there are $m$ active dimensions and $m > r$. From (\ref{eq:loss_ineq}), in this case $\tilde{r} = m - r$, and the loss is 

\begin{equation}
    \frac{1}{\gamma} \mathbb{E}_{q_{\phi_{\gamma}}(z|x)} [||x - \mu_x(z)||^2] + d \log (2\pi \gamma) - 2\sum_{i=1}^{r} \log \lambda_i - 2\sum_{i=r+1}^{r+\tilde{r}} \log \lambda_i + O(1)
\end{equation}

since here $\lim_{\gamma \rightarrow 0} \lambda_i=0$, $- 2\sum_{i=r+1}^{r+\tilde{r}} \log \lambda_i$ is monotonically increases as $\tilde{r}$ increases at the rate of $\Omega(\log \frac{1}{\gamma})$. For the reconstruction term, it is unaffected since we only use the first $r$ latent dimensions for reconstruction. Therefore, the loss will increase at the rate of $\Omega(\log \frac{1}{\gamma})$, which is larger than the loss when $m=r$.

In conclusion, only when the number of active dimensions equals $r$, and these active dimensions' encoder variance $\sigma_z^2(x) = O(\gamma)$ as $\gamma \rightarrow 0$, the optimal cost is $(d-r)\log \gamma + O(1)$.

\section{Proof of Theorem 2}

\paragraph{Summary of the proof} We focus on analyzing the loss conditioned on a specific $c$, which is defined as $\mathcal{L}_c(\theta, \phi)$. We then first construct the proof when $p_\theta(z|c) = p(z)$, i.e. a parameter free prior, and then extend to the case when the prior involves the conditioning variable. The logic flow is as follows:
\begin{enumerate}
    \item The prior is independent of $c$
    \begin{enumerate}
        \item Following the same proof idea as in Theorem~1, when the number of active dimensions whose encoder variance $\sigma^2_{z}(x, c; \phi) = O(\gamma)$ is less than $r-k$, where $k$ is the number of effective dimensions of $c$ used in the decoder, the reconstruction error will grow at a rate of $O(\frac{1}{\gamma})$. This is proven in Section \ref{ext_cvae};
        \item In Section \ref{std_bound} we show that both the upper bound and the lower bound are $(d-r+t)\log \gamma$ and the exact number of active dimensions is $r-t$.
    \end{enumerate}
    \item The prior is a function of $c$
    \begin{enumerate}
        \item Since involving $c$ in the prior will not affect the reconstruction term, we extend the conclusion in Section \ref{ext_cvae} to the general case;
        \item Show that both the upper bound and the lower bound are $(d-r+t)\log \gamma$ and the exact number of active dimensions is $r-t$. The proof is in Section \ref{param_bound}.
    \end{enumerate}
\end{enumerate}

Under CVAE setting, we first make some denotations for proof. Since the encoder, prior, and decoder share the same condition $c$, the model has flexibility to use part of each $c$ from the three networks. Denote $t$ as the number of effective dimensions of $c$, and $k$ as the number of effective dimensions of $c$ used in the decoder $p_\theta(x|z, c)$, i.e. there exists a pair of encoder and decoder, s.t. $\mu_x(c) = \varphi^{-1}(u_{1:k})$, where $0 \leq k \leq t$ and is a learnable parameter. The encoder and prior use the rest $t - k$ effective dimensions, i.e. $\mu_x(\mu_z(c)) = \varphi^{-1}(u_{k+1:t})$, and this part of information will be included in the latent variable $z$. 

\subsection{When the prior is independent of $c$, i.e. $p_\theta(z|c) = p(z)$}

In this case, we can write the cost as:

$$
\begin{aligned}
    2\mathcal{L}_c(\theta, \phi) =& 2\int_{\mathcal{X}_c} \{-\mathbb{E}_{q_\phi (z|x, c)}[\log p_\theta(x|z, c)] + \mathbb{KL}[q_\phi(z|x, c)||p(z)]\} \omega^c_{gt}(dx) \\
    =& \int_{\mathcal{X}_c} \frac{1}{\gamma} \mathbb{E}_{q_{\phi}(z|x, c)} [||x - \mu_x(z, c)||^2] + d \log (2\pi \gamma) + 2\mathbb{KL}(q_\phi(z|x, c)||p(z)) \omega^c_{gt}(dx) \\
    =& \frac{1}{\gamma} \int_{\mathcal{X}_c} \mathbb{E}_{q_{\phi_{\gamma}}(z|x, c)} [||\varphi^{-1}(u_{1:k}) - \mu_x(c)||^2 + ||\varphi^{-1}(u_{k+1: r}) - \mu_x(z_{k+1:r})||^2 + \\
     & d \log (2\pi \gamma) +2\mathbb{KL}(q_\phi(z|x, c)||p(z)) \omega^c_{gt}(dx)\\
    =&  \frac{1}{\gamma} \int_{\mathcal{X}_c} \mathbb{E}_{q_{\phi_{\gamma}}(z|x, c)} [||\varphi^{-1}(u_{k+1: r}) - \mu_x(z_{k+1:r})||^2 + d \log (2\pi \gamma) +2\mathbb{KL}(q_\phi(z|x, c)||p(z)) \omega^c_{gt}(dx)\\
\end{aligned}
$$

\subsubsection{The number of active dimensions whose encoder variance $\sigma^2_{z}(x, c; \phi) = O(\gamma)$ is less than $r-k$}
\label{ext_cvae}
Following the same proof idea in Theorem~1, assume there is no active dimension in $\sigma_z(x; \phi)$. For the reconstruction term, it is equivalent to reconstruct a $(r-k)$-dimensional manifold. Thus we can find an lower bound of the cost

\begin{equation}
    \begin{aligned}
        2\mathcal{L}_c(\theta, \phi) \geq& \frac{C'}{\gamma} + \int_{\mathcal{X}_c} \left[ d \log \gamma + 2\mathbb{KL}(q_\phi(z|x, c)||p(z)) + O(1) \right] \omega^c_{gt}(dx) \\
        = & \frac{C'}{\gamma} + \int_{\mathcal{X}_c} \left[ d \log \gamma - \log |\sigma^2_{z}(x, c; \phi)_{1: k}| - \log |\sigma^2_{z}(x, c; \phi)_{k+1:\kappa-t+k}| + O(1) \right] \omega^c_{gt}(dx)\\
    \end{aligned}
\end{equation}

where $C' = \int_{z_{1:r-k}} \min\{\frac{\alpha}{\sigma_l^{r-k}}e^{-\frac{1}{2}\sigma_l^{-2(r-k)}N} , \frac{\alpha'}{\sigma_u^{r-k}}e^{-\frac{r-k}{2}\sigma_u^{-2(r-k)+2}}\} dz_{1:r-k}$. 

The first term $\frac{C'}{\gamma}$ grows at a rate of O($\frac{1}{\gamma}$). Because $\sigma_z^2$ is at a lower rate than $\gamma$, we have $$O(d \log \gamma - \log |\sigma^2_z(x, c; \phi)|) < O(- (d - \kappa) \log \frac{1}{\gamma})$$ and the right part decreases at a rate of $\log \frac{1}{\gamma}$. When $\gamma \rightarrow 0$, $O(\frac{1}{\gamma}) > O(\log \frac{1}{\gamma})$, which means the increase from reconstruction term cannot be offset by the decrease from the KL term. Moreover, from the fact that $O(\frac{1}{\gamma}) > O(\log \frac{1}{\gamma})$, when $\gamma$ is small enough, the loss is monotonically decreasing with $\gamma$. 

Therefore, when $\gamma \rightarrow 0$, the lower bound cannot approach $-\infty$, which means at this case, the model can never achieve optimum. Thus, there must exist some active dimensions whose variance satisfies $\sigma^2_{z}(x, c; \phi)_i = O(\gamma), i=1, \dots, \kappa$ as $\gamma \rightarrow 0$ to reach the global optimum, and it is showed in $C'$ that as long as the number of such active dimensions exceeds $r-k$, as $\gamma$ approaches zero, the reconstruction term is at most at the rate of $O(1)$. Next, we will show that when there exist at least $r-k$ such active dimensions, the CVAE model's optimum is achievable.

\subsubsection{Bounds of CVAE cost}
\label{std_bound}
\paragraph{The upper bound}
We have $z_{1:t-k} = \mu_{z}(x, c; \phi)_{1: t-k} + \sigma_{z}(x, c; \phi)_{1:t-k}\varepsilon_1$ and $z_{t-k+1:r} = \mu_{z}(x, c; \phi)_{t-k+1:r-k} + \sigma_{z}(x; \phi)_{t-k+1:r-k}\varepsilon_2$, where $\varepsilon_1 \sim N(0, I^{t-k}), \varepsilon_2 \sim N(0, I^{r-t})$.

The loss is:

\begin{equation}
    \begin{aligned}
         &\frac{1}{\gamma} \mathbb{E}_{q_{\phi_{\gamma}}(z|x, c)} [||x - \mu_x(z, c)||^2] + d \log (2\pi \gamma) + 2\mathbb{KL}(q_\phi(z|x, c)||p(z))\\
         = & \frac{1}{\gamma} \mathbb{E}_{q_{\phi_{\gamma}}(z|x, c)} [||\varphi^{-1}(u_{1:k}) - \mu_x(c)||^2 + ||\varphi^{-1}(u_{k+1:t}) - \mu_x(z_{1:t-k})||^2 +  \\
         & ||\varphi^{-1}(u_{t+1:r}) - \mu_x(z_{t-k+1:r-k})||^2] + d \log (2\pi \gamma) +2\mathbb{KL}(q_\phi(z|x, c)||p(z))\\
         \leq & \frac{1}{\gamma} \mathbb{E}_{\varepsilon_1 \sim N(0, I^{t-k})}[||L \sigma_{z}(x, c; \phi)_{1:t-k} \varepsilon_1||^2] + \frac{1}{\gamma} \mathbb{E}_{\varepsilon_2 \sim N(0, I^{r-t})}[||L \sigma_{z}(x, c; \phi)_{t-k+1:r-k} \varepsilon_2||^2] + \\
         & d \log \gamma - \log |\sigma^2_{z}(x, c; \phi)_{1: t-k}| - \log |\sigma^2_z(x, c; \phi)_{t-k+1:r-k}| + O(1)\\
    \end{aligned}
\end{equation}

Denote $\sigma^2_{z}(x, c; \phi)_{1: t-k}$ as $\sigma^2_{z}(c, \phi)$ for simplicity, and denote the upper bound of loss as $\mathcal{L}^u_c$. Take the derivative of $\sigma_{z}(c; \phi)$ and $\sigma_{z}(x, c; \phi)_{t-k+1:r-k}$ separately. Because the diagonal elements in $\sigma_{z}(x, c; \phi)$ are independent, we can make both achieve optimum. We have:

\begin{equation}
    \begin{aligned}
         \mathcal{L}^u_c(\gamma, k) =& - (t-k) \log \gamma - (r-t) \log \gamma + d \log \gamma + O(1)\\
    \end{aligned}
\end{equation}

To minimize $\mathcal{L}^u_c(\gamma, k)$, the optimal $k$ is $t$, thus the upper bound is:
\begin{equation}
    \begin{aligned}
         \mathcal{L}^u_c(\gamma) =& (d - r + t) \log \gamma + O(1)\\
    \end{aligned}
\end{equation}

\paragraph{The lower bound} We have show that there must be at least $r-k$ active dimension at a rate of $O(\gamma)$, otherwise the loss will increase at a rate of $O(\gamma)$. We can get a lower bound

\begin{equation}
    \begin{aligned}
        &\frac{1}{\gamma} \mathbb{E}_{q_{\phi_{\gamma}}(z|x, c)} [||x - \mu_x(z, c)||^2] + d \log (2\pi \gamma) + 2\mathbb{KL}(q_\phi(z|x, c)||p(z))\\
        \geq & d \log \gamma - \log |\sigma_{z}^2(x, c; \phi)_{1:r-k}| - \log |\sigma_{z}^2(x, c; \phi)_{r-k+1:\kappa}| + O(1)\\
        \geq & d \log \gamma - (r-k) \log \gamma - \log |\sigma_{z}^2(x, c; \phi)_{r-k+1:\kappa}| + O(1)\\
        \geq & (d - r + k) \log \gamma + O(1)
    \end{aligned}
\end{equation}

Denote the lower bound as $\mathcal{L}^l_c(\gamma, k)$, to minimize it, we have $k=t$, thus the lower bound is 

$$\mathcal{L}_c^l(\gamma) = (d-r+t) \log \gamma + O(1)$$

Both $\mathcal{L}_c^u$ and $\mathcal{L}_c^l$ are at a rate of $O(\log \gamma)$, we come to the conclusion that the ELBO is $(d - r +t)\log \gamma + O(1)$ and the number of active dimensions is $r-t$ when $p_\theta(z|c) = p(z)$.

\subsection{The general case}
\label{param_bound}
Define a trainable parametric prior of $z$, i.e. $z \sim N(\mu_{z}(c; \theta), \sigma^2_{z}(c; \theta))$. Since involving $c$ in the prior doesn't affect the reconstruction term, we have the conclusion in Section~\ref{ext_cvae} that there are at least $r - k$ active latent dimensions at a rate of $O(\gamma)$. Without loss of generality, we assume the first $r-k$ dimension of $\sigma_{z}^2(x, c; \phi)$, i.e. $\sigma_{z}^2(x, c; \phi)_{1:r-k} = O(\gamma)$.

\paragraph{The upper bound } The loss is:

\begin{equation}
    \begin{aligned}
         &\frac{1}{\gamma} \mathbb{E}_{q_{\phi_{\gamma}}(z|x, c)} [||x - \mu_x(z, c)||^2] + d \log (2\pi \gamma) + 2\mathbb{KL}(q_\phi(z|x, c)||p(z|c))\\
         \leq & \frac{1}{\gamma} \mathbb{E}_{\varepsilon_1 \sim N(0, I^{t-k})}[||L \sigma_{z}(c; \phi) \varepsilon_1||^2] + \frac{1}{\gamma} \mathbb{E}_{\varepsilon_2 \sim N(0, I^{r-k})}[||L \sigma_{z}(x, c; \phi)_{t-k+1:r-k} \varepsilon_2||^2] + d \log (2\pi \gamma) -\\
         & \log |\sigma^2_{z}(c, \phi)| - \log |\sigma^2_{z}(x, c; \phi)_{t-k+1:r-k}| - \log |\sigma^2_{z}(x, c; \phi)_{r-k+1:\kappa}|  + \log |\sigma_{z}^2(c; \theta)_{1:t-k}| + \log |\sigma_{z}^2(c; \theta)_{t-k+1:\kappa}|\\
         & + (\mu_{{z_\phi}(1:t-k)}- \mu_{{z_\theta}(1:t-k)})^T \sigma^2_{z}(c; \theta)_{1:t-k}^{-1} (\mu_{{z_\phi}(1:t-k)}- \mu_{{z_\theta}(1:t-k)}) \\
         & + (\mu_{{z_\phi}(t-k+1:\kappa)}- \mu_{{z_\theta}(t-k+1:\kappa)})^T \sigma^2_{z}(c; \theta)_{t-k+1:\kappa}^{-1} (\mu_{{z_\phi}(t-k+1:\kappa)}- \mu_{{z_\theta}(t-k+1:\kappa)})\\
         & - \kappa + tr(\sigma^2_{z}(c, \phi)/\sigma^2_{z}(c; \theta)_{1:t-k}) + tr(\sigma^2_{z}(x; \phi)/\sigma^2_{z}(c; \theta)_{t-k+1:\kappa})\\
    \end{aligned}
\end{equation}

Since we can only control $k$ dimensions of the prior when training, take the derivative of $\mu_{z}(c; \theta)_{1:k}$ and $\sigma_{z}(c; \theta)_{1:k}$, we have

\begin{equation}
    \begin{aligned}
         \mu_{z}(c; \theta)_{1:t-k}^* &= \mu_{z}(c; \phi)\\
         \sigma_{z}^2(c; \theta)_{1:t-k}^* &= (\mu_{z}(c; \phi) - \mu_{z}(c; \theta)_{1:t-k}^*)(\mu_{z}(c; \phi) - \mu_{z}(c; \theta)_{1:t-k}^*)^T + \sigma^2_{z}(c, \phi) = \sigma^2_{z}(c, \phi)
    \end{aligned}
\end{equation}

Let them achieve the optimal values. The loss becomes

\begin{equation}
\label{eq:optim_loss_prior}
    \begin{aligned}
        & \frac{1}{\gamma} \mathbb{E}_{\varepsilon_1 \sim N(0, I^{t-k})}[||L \sigma_{z}(c; \phi) \varepsilon_1||^2] + \frac{1}{\gamma} \mathbb{E}_{\varepsilon_2 \sim N (0, I^{r-k})}[||L \sigma_{z}(x, c; \phi)_{t-k+1:r-k} \varepsilon_2||^2] + d \log(2 \pi \gamma)\\
        & - \log |\sigma^2_{z}(x, c; \phi)_{t-k+1:r-k}| - \log |\sigma^2_{z}(x, c; \phi)_{r-k+1:\kappa}| + \log |\sigma_{z}^2(c; \theta)_{r-k+1:\kappa}| + tr(\sigma^2_{z}(x; \phi)/\sigma^2_{z}(c; \theta)_{k+1:\kappa}) \\
        & + (\mu_{{z_\phi}(t-k+1:\kappa)}- \mu_{{z_\theta}(t-k+1:\kappa)})^T \sigma^2_{z}(c; \theta)_{t-k+1:\kappa}^{-1} (\mu_{{z_\phi}(t-k+1:\kappa)}- \mu_{{z_\theta}(t-k+1:\kappa)}) + t - k - \kappa
    \end{aligned}
\end{equation}

From (\ref{eq:optim_loss_prior}) we observe that if we have a flexible enough prior, there are $t-k$ latent dimensions that won't provide any loss both in reconstruction and kl term. To minimize (\ref{eq:optim_loss_prior}), $\sigma^2_{z}(c, \phi)^* = 0$ and $\sigma_{z}^2(x; \phi)^*_{t-k+1:r-k} = \gamma \frac{I}{L^2}$. Let them be the optimums, and view the terms that are irrelevant with $\gamma$ when it approaches 0 as constants, we have

\begin{equation}
    \begin{aligned}
        \mathcal{L}_c^{u'} = (d - r + t) \log \gamma + O(1)
    \end{aligned}
\end{equation}

\paragraph{The lower bound} to get the lower bound, we have

\begin{equation}
    \begin{aligned}
     &\frac{1}{\gamma} \mathbb{E}_{q_{\phi_{\gamma}}(z|x, c)} [||x - \mu_x(z, c)||^2] + d \log (2\pi \gamma) + 2\mathbb{KL}(q_\phi(z|x, c)||p(z|c))\\
     \geq& d \log \gamma - \log |\sigma^2_{z}(x, c; \phi)_{t-k+1:r-k}| - \log |\sigma^2_{z}(x, c; \phi)_{r-k+1:\kappa}| + \log |\sigma_{z}^2(c; \theta)_{r-k+1:\kappa}| + \\
        & tr(\sigma^2_{z}(x; \phi)/\sigma^2_{z}(c; \theta)_{k+1:\kappa})  + (\mu_{{z_\phi}(t-k+1:\kappa)}- \mu_{{z_\theta}(t-k+1:\kappa)})^T \sigma^2_{z}(c; \theta)_{t-k+1:\kappa}^{-1} (\mu_{{z_\phi}(t-k+1:\kappa)}- \mu_{{z_\theta}(t-k+1:\kappa)}) + O(1)\\
    =& d \log \gamma - \log |\sigma^2_{z}(x, c; \phi)_{t-k+1:r-k}| +O(1)\\
    \geq & d \log \gamma - (r - t)\log \gamma + O(1)\\
    =& (d-r+t)\log \gamma + O(1)
    \end{aligned}
\end{equation}

The last inequality comes from the conclusion that there are at least $r-k$ active dimensions at a rate of $O(\gamma)$ and the loss is monotonously increase with $\gamma$. In this case $k$ can be any integer in $[0, t]$, thus we cannot determine how many dimensions are used by the encoder and decoder separately. But no matter what value $k$ is, the cost of CVAE is 

$$
(d-r+t)\log \gamma + O(1)
$$

In conclusion, after integrating over $\mathcal{C}$, we have $$\mathcal{L}(\theta^\ast, \phi^\ast) = \int_\mathcal{C} \mathcal{L}_c(\theta^\ast, \phi^\ast) \nu_{gt}(dc) = (d-r+t)\log \gamma + O(1)$$

\section{Proof of Theorem 3}

\paragraph{Summary of the proof} We first define a space of sequences, and then separate the sequences into two categories according to the performance of the KL term. In Section \ref{thm3_kl1}, we analyze the case when the Kl term equals $O(\log \frac{1}{\gamma})$, and in Section \ref{thm3_kl2}, the rate of KL term is higher than $O(\log \frac{1}{\gamma})$. In both categories, we prove that the whole cost cannot go to $-\infty$.

Let $\theta^\ast, \phi^\ast = \arg \min_{\theta, \phi}\mathcal{L}(\theta, \phi)$. Define $S \subset \mathcal{X}$ as the set of the sequences, and the sequence is defined as $\{x_l\}_{l=1}^\infty \in S$.

Consider when $l$ equals to a constant $l_0$, we have the prior as $q_{\phi^\ast}(z|x_{<l_0})$, and encoder as $q_{\phi^\ast}(z|x_{\leq l_0})$. Next, consider $l=l_0+1$, we have the prior as $q_{\phi^\ast}(z|x_{\leq l_0})$, which is exactly the same as the encoder at $l=l_0$, and the encoder as $q_{\phi^\ast}(z|x_{\leq l_0+1})$. The cost function at these two points are

$$\mathcal{L}^{(l_0)}_c(\theta^\ast, \phi^\ast) = -\mathbb{E}_{q_{\phi^\ast} (z|x_{\leq l_0})}[\log p_{\theta^\ast}(x_{l_0}|z, x_{< l_0})] + \mathbb{KL}[q_{\phi^\ast}(z|x_{\leq l_0})||q_{\phi^\ast}(z| x_{< l_0})]$$

and 

$$\mathcal{L}^{(l_0+1)}_c(\theta^\ast, \phi^\ast) = -\mathbb{E}_{q_{\phi^\ast} (z|x_{\leq l_0+1})}[\log p_{\theta^\ast}(x_{l_0+1}|z, x_{\leq l_0})] + \mathbb{KL}[q_{\phi^\ast}(z|x_{\leq l_0+1})||q_{\phi^\ast}(z| x_{\leq l_0})]$$

respectively.

Next, we separate the sequences with varied values into \textbf{two} cases.

\subsection{KL term is at a rate of $O(\log \frac{1}{\gamma})$ when $\gamma \rightarrow 0$.}
\label{thm3_kl1}

Denote $\log q_\phi(z|x_{\leq l}) - \log q_\phi(z|x_{< l}) = f_{l}(\gamma) = O(\log \frac{1}{\gamma})$. In this setting, we analyze the reconstruction term

\begin{equation}
    \begin{split}
        &-\mathbb{E}_{q_{\phi^\ast} (z|x_{\leq l_0})}[\log p_{\theta^\ast}(x_{l_0}|z, x_{< l_0})] \\
        =&\int_\mathcal{Z} q_{\phi^\ast} (z|x_{\leq l_0}) \log p_{\theta^\ast}(x_{l_0}|z, x_{< l_0}) dz\\
        =&\frac{1}{\gamma} \int_{\mathcal{Z}} q_{\phi^\ast}(z|x_{\leq l_0}) [||x_{l_0} - \mu_x^\ast(z)||^2] dz + d\log (2\pi \gamma)
    \end{split}
\end{equation}

Similarly we have 

\begin{equation}
\begin{split}
    &-\mathbb{E}_{q_{\phi^\ast} (z|x_{\leq l_0+1})}[\log p_{\theta^\ast}(x_{l_0+1}|z, x_{\leq l_0})] \\
    =&\frac{1}{\gamma} \int_{\mathcal{Z}} q_{\phi^\ast}(z|x_{\leq l_0+1}) [||x_{l_0+1} - \mu_x^\ast(z)||^2] dz + d\log (2\pi \gamma)
\end{split}
\end{equation}

With the condition that $\log q_\phi(z|x_{\leq l}) - \log q_\phi(z|x_{< l}) = f_{l}(\gamma) = O(\log \frac{1}{\gamma})$, we have

\begin{equation}
\begin{split}
    & \mathbb{KL}[q_\phi(z|x_{\leq l})||q_\phi(z| x_{< l})]\\
    =& \mathbb{E}_{q_\phi(z|x_{\leq l})} \left[ \log q_\phi(z|x_{\leq l}) - \log q_\phi(z|x_{< l})\right] \\
    \leq& \mathbb{E}_{q_\phi(z|x_{\leq l})} [f_{l}(\gamma)] = f_{l}(\gamma)
\end{split}
\end{equation}

That shows KL term is either small than a constant or goes to infinity at a slower rate than rate of $\log \frac{1}{\gamma}$ when $\gamma \rightarrow 0$. We can also get $q_\phi(z|x_{< l}) \geq \frac{1}{e^{f_{l}(\gamma)}} q_\phi(z|x_{< l})$, from which we have

\begin{equation}
    \begin{split}
        q_\phi(z|x_{< l}) - q_\phi(z|x_{\leq l}) \geq& q_\phi(z|x_{\leq l})(\frac{1}{e^{f_{l}(\gamma)}} - 1)
    \end{split}
\end{equation}

Together we have

\begin{equation}
    \begin{split}
        &-\mathbb{E}_{q_{\phi^\ast} (z|x_{\leq l_0})}[\log p_{\theta^\ast}(x_{l_0}|z, x_{< l_0})] -\mathbb{E}_{q_{\phi^\ast} (z|x_{\leq l_0+1})}[\log p_{\theta^\ast}(x_{l_0+1}|z, x_{\leq l_0})]\\
        =& \frac{1}{\gamma} \left[\int_{\mathcal{Z}} q_{\phi^\ast}(z|x_{\leq l_0}) ||x_{l_0} - \mu_x^\ast(z)||^2 dz + \int_{\mathcal{Z}} q_{\phi^\ast}(z|x_{\leq l_0 + 1}) ||x_{l_0+1} - \mu_x^\ast(z)||^2 dz \right]  + 2d\log (2\pi \gamma)\\
        =& \frac{1}{\gamma} [\int_{\mathcal{Z}} q_{\phi^\ast}(z|x_{\leq l_0 + 1}) \left[||x_{l_0} - \mu_x^\ast(z)||^2 + ||x_{l_0 + 1} - \mu_x^\ast(z)||^2\right] dz + \\
        &\int_{\mathcal{Z}} \left[q_{\phi^\ast}(z|x_{\leq l_0}) - q_{\phi^\ast}(z|x_{\leq l_0+1}) \right] ||x_{l_0} - \mu_x^\ast(z)||^2 dz ]  + 2d\log (2\pi \gamma)\\
        &\int_{\mathcal{Z}} \left[q_{\phi^\ast}(z|x_{\leq l_0}) - q_{\phi^\ast}(z|x_{\leq l_0+1}) \right] ||x_{l_0} - \mu_x^\ast(z)||^2 dz ]  + 2d\log (2\pi \gamma)\\
        \geq& \frac{1}{\gamma} [\int_{\mathcal{Z}} q_{\phi^\ast}(z|x_{\leq l_0 + 1}) \left[||x_{l_0} - \mu_x^\ast(z)||^2 + ||x_{l_0 + 1} - \mu_x^\ast(z)||^2\right] dz + \\
        & (\frac{1}{e^{f_{l_0}(\gamma)}} - 1) \int_{\mathcal{Z}} q_\phi(z|x_{\leq l_0 + 1}) ||x_{l_0} - \mu_x^\ast(z)||^2] dz + 2d\log (2\pi \gamma)\\
        =& \frac{1}{\gamma} \int_{\mathcal{Z}} q_{\phi^\ast}(z|x_{\leq l_0 + 1}) \left[\frac{1}{e^{f_{l_0}(\gamma)}}||x_{l_0} - \mu_x^\ast(z)||^2 + ||x_{l_0 + 1} - \mu_x^\ast(z)||^2\right] dz + 2d\log (2\pi \gamma)
    \end{split}
\end{equation}

For any $l_0 = 1, 2, \dots$ and all $z \in \mathcal{Z}$, we have the following cases:
    \begin{enumerate}
    \item \textbf{For any $z \in \mathcal{Z}_1$, $\mu_x^\ast(z) = x_{l_0}$ and $\mu_x^\ast(z) \neq x_{l_0+1}$}. We have $\frac{||x_{l_0} - \mu_x^\ast(z)||^2}{\gamma} \rightarrow \infty$ at a rate of $O(\frac{1}{\gamma})$. 
    \item \textbf{For any $z \in \mathcal{Z}_2$,  $\mu_x^\ast(z) = x_{l_0+1}$ and $\mu_x^\ast(z) \neq x_{l_0}$}. We have $\frac{||x_{l_0+1} - \mu_x^\ast(z)||^2}{\gamma e^{f_{l_0}(\gamma)}} \rightarrow \infty$ at a rate of $O(\frac{1}{\gamma e^{f_{l_0}(\gamma)}})$. 
    \item \textbf{For any $z \in \mathcal{Z}_3$, $\mu_x^\ast(z) \neq x_{l_0}$ and $\mu_x^\ast(z) \neq x_{l_0+1}$.} Both cases above cause the norm term equal $\Omega(1)$.
    \end{enumerate}

With the setting, the lower bound of reconstruction term is

\begin{equation}\label{thm3:lower_bound}
    \begin{split}
        & \frac{1}{\gamma} \int_{\mathcal{Z}} q_{\phi^\ast}(z|x_{\leq l_0 + 1}) \left[\frac{1}{e^{f_{l_0}(\gamma)}}||x_{l_0} - \mu_x^\ast(z)||^2 + ||x_{l_0 + 1} - \mu_x^\ast(z)||^2\right] dz + 2d\log (2\pi \gamma) \\
        =&  \frac{1}{\gamma} \int_{\mathcal{Z}_1} q_{\phi^\ast}(z|x_{\leq l_0 + 1}) ||x_{l_0 + 1} - \mu_x^\ast(z)||^2 dz + \frac{1}{\gamma e^{f_{l_0}(\gamma)}} \int_{\mathcal{Z}_2} q_{\phi^\ast}(z|x_{\leq l_0 + 1}) ||x_{l_0} - \mu_x^\ast(z)||^2  dz + \\
        & \frac{1}{\gamma} \int_{\mathcal{Z}_3} q_{\phi^\ast}(z|x_{\leq l_0 + 1}) \left[\frac{1}{e^{f_{l_0}(\gamma)}}||x_{l_0} - \mu_x^\ast(z)||^2 + ||x_{l_0 + 1} - \mu_x^\ast(z)||^2\right] dz + 2d\log (2\pi \gamma)\\
        \geq& \frac{1}{\gamma} \int_{\mathcal{Z}_1 \cup \mathcal{Z}_3} q_{\phi^\ast}(z|x_{\leq l_0 + 1}) ||x_{l_0 + 1} - \mu_x^\ast(z)||^2 dz + 2d\log (2\pi \gamma)
    \end{split}
\end{equation}

Since the probability mass of $x_l$ conditioned on $x_{<l}$ lies on a manifold with at least 1 dimension, i.e. we exclude deterministic sequences, there must exist a sequence $\{x_l\}^{i_0}$, in which $\sum_{l=1}^\infty \int_\mathcal{Z} q_{\phi^\ast}(z|x_{\leq l + 1}) ||x_{l + 1} - \mu_x^\ast(z)||^2 dz \geq C$, where $C$ is a constant, otherwise all the sequences $\{x_l\}^i \in S, i=1, 2, \dots$ share the same values which violates our assumption.

Then for (\ref{thm3:lower_bound}), there must exist a constant $C'$, such that $$\int_{\mathcal{Z}_1 \cup \mathcal{Z}_3} q_{\phi^\ast}(z|x_{\leq l_0 + 1}) ||x_{l_0 + 1} - \mu_x^\ast(z)||^2 dz \geq C'$$ Thus, the lower bound of the cost is $$\frac{C'}{\gamma} - 2d\log \frac{1}{2 \pi \gamma}$$When $\gamma$ goes to zero, $O(\frac{1}{\gamma}) > O(\log \frac{1}{\gamma})$. We get the conclusion that $\mathcal{L}_c(\theta, \phi) = \int_\mathcal{X} \Omega(1) \omega_{gt}(dx)$ for any $\theta$ and $\phi$.

\subsection{KL term goes to infinity at a rate higher than $O(\log \frac{1}{\gamma})$.}  
\label{thm3_kl2}
In this case, we have 

\begin{equation}
\begin{split}
    & 2\mathbb{KL}[q_{\phi^\ast}(z|x_{\leq l_0+1})||q_{\phi^\ast}(z| x_{\leq l_0})]\\
    =& \log \frac{|\sigma_z^2(x_{\leq l_0})|}{|\sigma_z^2(x_{\leq l_0+1})|} - \kappa + (\mu_z(x_{\leq l_0+1}) - \mu_z(x_{\leq l_0}))^T\sigma_z^{-2}(x_{\leq l_0})(\mu_z(x_{\leq l_0+1}) - \mu_z(x_{\leq l_0})) \\
    &+ tr(\sigma_z^{-2}(x_{\leq l_0})\sigma_z^{2}(x_{\leq l_0+1}))
\end{split}
\end{equation}

Thus it can only happen when there are some dimensions where $\sigma^2_z(x_{\leq l_0})$ is active while $\sigma^2_z(x_{\leq l_0+1})$ is not, which indicate that $tr(\sigma_z^{-2}(x_{\leq l_0})\sigma_z^{2}(x_{\leq l_0+1})) \rightarrow \infty$ at a rate of $\Omega(\frac{1}{\gamma})$. We have 

\begin{equation}
    \begin{split}
       & -2\mathbb{E}_{q_{\phi^\ast} (z|x_{\leq l_0+1})}[\log p_{\theta^\ast}(x_{l_0+1}|z, x_{\leq l_0})] + 2\mathbb{KL}[q_{\phi^\ast}(z|x_{\leq l_0+1})||q_{\phi^\ast}(z| x_{\leq l_0})]\\
       \geq& d\log (2\pi \gamma) + tr(\sigma_z^{-2}(x_{\leq l_0})\sigma_z^{2}(x_{\leq l_0+1})) + \log \frac{|\sigma_z^2(x_{\leq l_0})|}{|\sigma_z^2(x_{\leq l_0+1})|}\\
       & - \kappa + (\mu_z(x_{\leq l_0+1}) - \mu_z(x_{\leq l_0}))^T\sigma_z^{-2}(x_{\leq l_0})(\mu_z(x_{\leq l_0+1}) - \mu_z(x_{\leq l_0}))
    \end{split}
\end{equation}

Because $d\log (2\pi \gamma) + \log \frac{|\sigma_z^2(x_{\leq l_0})|}{|\sigma_z^2(x_{\leq l_0+1})|} \rightarrow -\infty$ at a rate of $O(\log \gamma)$ while $tr(\sigma_z^{-2}(x_{\leq l_0})\sigma_z^{2}(x_{\leq l_0+1})) \rightarrow \infty$ at a rate of $\Omega(\frac{1}{\gamma})$, the whole loss will go to infinity.

In summary, in both cases, when summing over $l$, $\mathcal{L}(\theta, \phi)$ will go to infinity.

\section{Justification of Remark 1}

Consider a $\kappa$-simple CVAE model with encoder $q_\phi(z|x, c)$, prior $p_\theta(z|c)$ and decoder $p_\theta(x|z, c)$. Further, let $\mu_q(x, c), \sigma_q(x, c)$ be the distributional parameters for $z\sim q_\phi(z|x, c)$, $\mu_p(c), \sigma_p(c)$ be the distributional parameters for $z\sim p_\theta(z|c)$. Name this model as $M$, and we have its cost with regard to $(\theta, \phi)$ being

\begin{equation}
\begin{aligned}
    2\mathcal{L}_c(M; \theta, \phi) =& \int_\mathcal{X} \{-2\mathbb{E}_{q_\phi (z|x, c)}[\log p_\theta(x|z, c)] + 2\mathbb{KL}[q_\phi(z|x, c)||p_\theta(z| c)]\} \omega^c_{gt}(dx) \\
    =& \frac{1}{\gamma}\int_\mathcal{X} \int_\mathcal{Z} \mathcal{N}(z;\mu_q(x, c), \sigma_q(x,c))||x-\mu_x(z)||^2 dz \omega^c_{gt}(dx)\\
    &+\log(2\pi \gamma)  + \int_\mathcal{X}  2\mathbb{KL}[q_\phi(z|x, c)||p_\theta(z| c)] \omega^c_{gt}(dx)\\
    =& \frac{1}{\gamma}\int_\mathcal{X} \int_\mathcal{Z} \frac{1}{\sqrt{(2\pi\gamma)^d}}\text{exp}\{-\frac{||z-\mu_q||^2}{2 \sigma_q^2}\}||x-\mu_x(z)||^2 dz \omega_{gt}(dx)\\
    &+\log(2\pi \gamma) + \int_\mathcal{X} [\log \sigma_p - \log \sigma_q - \kappa + ||\mu_q - \mu_p||^2 / \sigma_p + \text{tr}(\sigma_q / \sigma_p)] \omega^c_{gt}(dx)
\end{aligned}
\end{equation}

Next, we construct another $\kappa$-simple CVAE, $M^\prime$, with a standard Gaussian prior, only using computation modules in $M$. 
Specifically, the new prior, decoder, and encoder are defined as:

\begin{itemize}
    \item Prior: $p^\prime(z^\prime)=\mathcal{N}(0, \text{I})$
    \item Decoder: $p^\prime(x|z^\prime, c) = p_\theta(x|z^\prime * \sigma_p(c) + \mu_p(c), c)$
    \item Encoder: $q^\prime(z|x, c)=\mathcal{N}(\mu_q^\prime, \sigma_q^\prime)$, where
    \begin{itemize}
        \item $\mu_q^\prime = (\mu_q(x, c) - \mu_p(c)) / \sigma_p(c)$
        \item $\sigma_q^\prime = \sigma_q(x, c) / \sigma_p(c)$
    \end{itemize}
\end{itemize}

With $M^\prime$ defined, we are going to show that it has the exact same cost value as the above one during training, i.e. $\mathcal{L}(M^\prime;\theta, \phi) = \mathcal{L}(M;\theta, \phi)$, and the generated data distribution during generation, i.e. $p^\prime(x|z^\prime, c)\mathcal{N}(z^\prime; 0, \text{I})\equiv p_\theta(x|z, c)p_\theta(z|c)$.

During training, we have $z^\prime\sim\mathcal{N}(\mu_q^\prime, \sigma_q^\prime)$, thus $z^\prime * \sigma_p(c) + \mu_p(c) \sim \mathcal{N}((\mu_q(x, c) - \mu_p(c)) / \sigma_p(c) * \sigma_p(c) + \mu_p(c), \sigma_q(x, c) / \sigma_p(c) * \sigma_p(c)) = \mathcal{N}(\mu_q(x, c), \sigma_q(x, c))$. Thus, we have

\begin{equation}
\begin{split}
    &\mathbb{E}_{q(z^\prime|x, c)}[\log p_\theta(x|z^\prime * \sigma_p(c) + \mu_p(c), c)]\\
    =&\frac{1}{\gamma} \int_\mathcal{Z} \mathcal{N}(z^\prime;\mu^\prime, \sigma^\prime)||x-\mu_x(z^\prime*\sigma_p(c) + \mu_p(c))||^2 dz^\prime  +\log(2\pi \gamma)\\
    =&\frac{1}{\gamma} \int_\mathcal{Z} \mathcal{N}(z;\mu_q(x, c), \sigma_q(x, c))||x-\mu_x(z)||^2 dz +\log(2\pi \gamma)\\
    =&\mathbb{E}_{q_\phi (z|x, c)}[\log p_\theta(x|z, c)]
\end{split}
\end{equation}


Besides, we also have

\begin{equation}
    \begin{split}
        &\mathbb{KL}[q^\prime(z^\prime|x, c)||\mathcal{N}(0, \text{I})] \\
        =&\mathbb{KL}[\mathcal{N}(\mu_q(x, c) - \mu_p(c)) / \sigma_p, \sigma_q(x, c) / \sigma_p(c)||\mathcal{N}(0, \text{I})]\\
        =&\frac{1}{2}[||\mu_q(x, c) - \mu_p(c)) / \sigma_p||^2 + \text{tr}(\sigma_q / \sigma_p) - \kappa - \log (\sigma_q/\sigma_p)] \\
        =&\frac{1}{2}[\log \sigma_p - \log \sigma_q - \kappa + ||\mu_q - \mu_p||^2 / \sigma_p + \text{tr}(\sigma_q / \sigma_p)]\\
        =&\mathbb{KL}[q_\phi(z|x, c)||p_\theta(z| c)]
    \end{split}
\end{equation}

In terms of generation equivalence, for any $z_p^\prime\sim \mathcal{N}(0, \text{I})$, we have

\begin{equation}
    \begin{split}
        &p^\prime(x|z^\prime, c)p(z^\prime; 0, \text{I})\\
        =&p_\theta(x|z^\prime * \sigma_p(c) + \mu_p(c))p(z^\prime; 0, \text{I})\\
        =&p_\theta(x|z)p(z; \mu_p(c), \sigma_p(c))\\
        =&p_\theta(x|z)p_\theta(z|c)
    \end{split}
\end{equation}

Therefore we conclude that $M'$ and $M$ share the same cost value, i.e. $\mathcal{L}_c(M^\prime; \theta, \phi) = \mathcal{L}_c(M; \theta, \phi)$, and equivalent data generation distributions.
\end{document}